\documentclass[journal]{IEEEtran}
\usepackage{booktabs}
\usepackage{amsmath}
\usepackage{array}   % 对齐
\usepackage[caption=false,font=footnotesize]{subfig} % 子图
\usepackage{url}

\usepackage{cite}
\usepackage{algorithm}  %%%
\usepackage{algorithmicx}
\usepackage{algpseudocode}
\usepackage{cleveref}
\usepackage{color}
\usepackage{graphicx}
\usepackage{float}
\usepackage{subfig}
\usepackage{caption,amssymb}

\begin{document}
\title{Multi-view Clustering via Deep Matrix Factorization and Partition Alignment}

\author{Chen~Zhang$^{*}$,~Siwei~Wang$^{*}$,~Jiyuan~Liu,~Sihang~Zhou,~Pei~Zhang, ~Xinwang~Liu$^{\dagger}$,~En~Zhu and ~Changwang~Zhang~$^{\dagger}$}
\maketitle

\begin{abstract}
Multi-view clustering (MVC) has been extensively studied to collect multiple source information in recent years. One typical type of MVC methods is based on matrix factorization to effectively perform dimension reduction and clustering. However, the existing approaches can be further improved with following considerations: $i)$ The current one-layer matrix factorization framework cannot fully exploit the useful data representations. $ii)$ Most algorithms only focus on the shared information while ignore the view-specific structure leading to suboptimal solutions. $iii)$ The partition level information has not been utilized in existing work. To solve the above issues, we propose a novel multi-view clustering algorithm via deep matrix decomposition and partition alignment. To be specific, the partition representations of each view are obtained through deep matrix decomposition, and then are jointly utilized with the optimal partition representation for fusing multi-view information. Finally, an alternating optimization algorithm is developed to solve the optimization problem with proven convergence. The comprehensive experimental results conducted on six benchmark multi-view datasets clearly demonstrates the effectiveness of the proposed algorithm against the SOTA methods.
\end{abstract}


\IEEEpeerreviewmaketitle

\begin{figure*}
	\includegraphics[width=\textwidth]{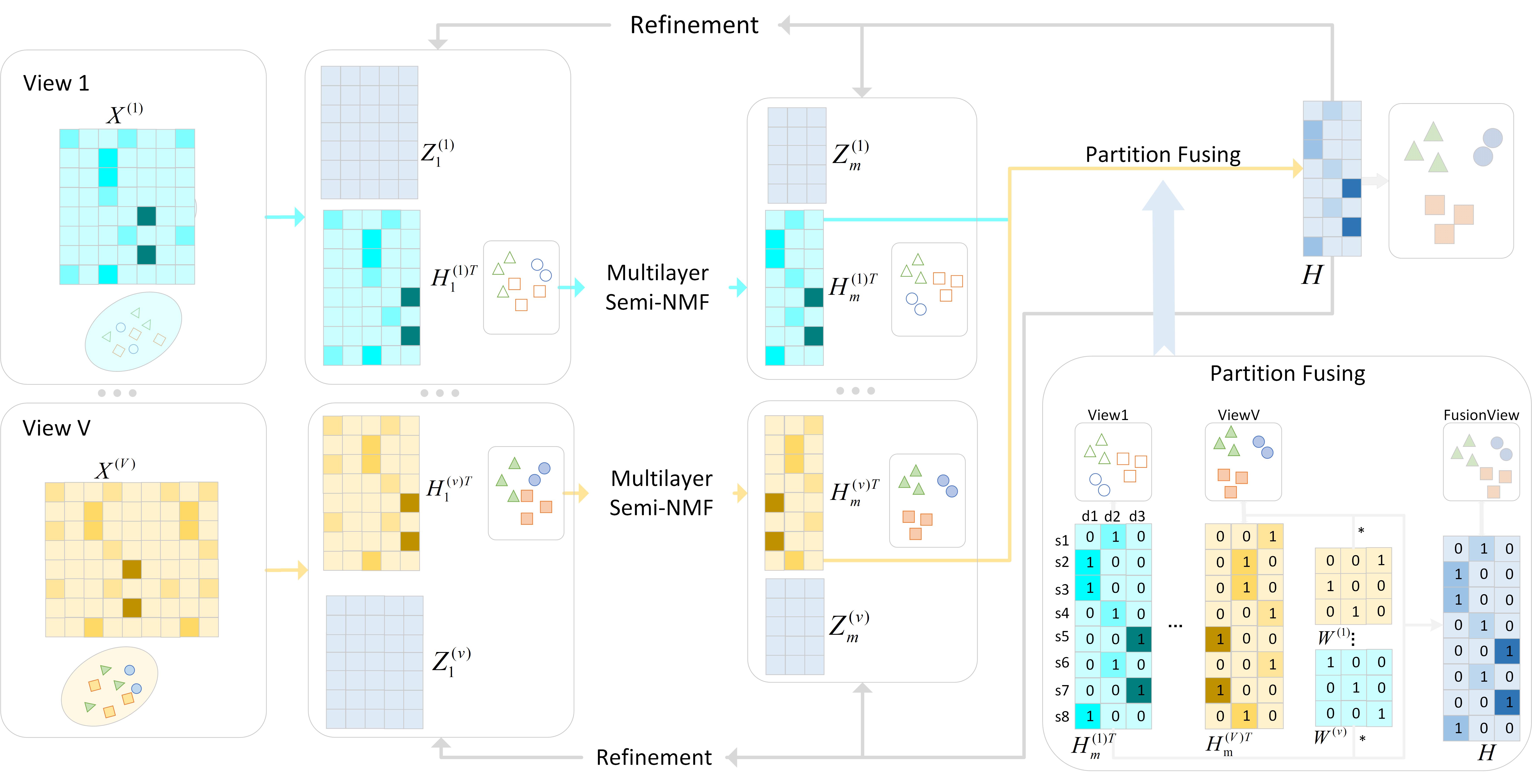}
	\caption{The illustration of our proposed MVC-DMF-PA. Multiple base partition matrix are obtained by deep semi-NMF firstly. Then a consensus partition matrix is learned by maximizing the alignment of this matrix with a uniformly weighted base partition matrix via optimal permutations. Finally, the deep matrix decomposition stage and late fusion stage are alternately boosted until convergence.}
	\label{fig:framework}
\end{figure*}

\section{Introduction}\label{intro}

In recent years, enormous data are collected from multiple sources or described by diverse attributes, which is known as multi-view data in most literature. For instance, an item can be represented with image illustration and short text description; person identify contains face image and voice information. With a large amount of unlabelled multi-view data, multi-view clustering is proposed to make full use of provided information and therefore has been attracted great attention. Existing multi-view clustering algorithms can be further classified into four categories by means of applied models: Co-training \cite{kumar_co-training_nodate, kumar_co-regularized_nodate, tan2020unsupervised}, multi-kernel learning \cite{li_multiple_nodate, huang2020robust, wang2020kernelized, chen2019jointly}, graph clustering \cite{wang2019multi,zhang_multilevel_2021} and subspace clustering \cite{chen_relaxed_2021,zhou_subspace_2020,wang2020learning}. The basic idea of early fusion is to fuse multiple feature or graph structure representations of multi-view into a single representation, after which a known single-view clustering algorithm can be applied. For example, graph-based clustering approach \cite{zhou2007spectral} constructs sample similarities under each view with graph structure and then fuse these graphs by using random walk strategy. Multi-kernel learning methods fuse multiple base kernels by linear or nonlinear combinations to obtain the optimal kernel for clustering. The subspace clustering \cite{kang2020partition} aims to find suitable low-dimensional representation and structure for each view, and then fuses them into a representation containing rich information for clustering. On the other hand, the approaches of late fusion ensemble the clustering results of individual views. Late fusion can be divided into integrated learning and collaborative training. The input to the integrated clustering algorithm is the result of clustering corresponding to multiple views. In \cite{bickel2004multi},  a consensus loss function for the distance between the final clustering results and the input clustering results is defined to obtain the clustering result. The focus of co-training is on how to get better clustering results during the co-train training process. \cite{kumar_co-training_nodate} obtains multiple clustering results by performing spectral embedding on each view and the obtained clustering results are used to influence the original representation of the other views. Moreover, \cite{wang2019multi} applies late fusion for multi-kernel k-means clustering and reduces the complexity of the algorithm and the time cost. Our proposed method belongs to Non-negative Matrix Factorization(NMF) clustering in subspace clustering and also to late fusion clustering.
% 最新的NMF

NMF is widely used in clustering because of its ability to handle high-dimensional data and to capture the underlying representation of different views. Some work \cite{ding2006orthogonal,zong2018multi,wang2017diverse} reduces the redundancy between different view representation by defining the diversity of views. The method in \cite{yang2020uniform} tends to generate distributions with uniform decomposition, making the learned representations more discriminative. Furthermore, cross-entropy cost function \cite{liu2020multi} and neighbor information \cite{chen2019multiview} are introduced to guide the learning process. Although NMF can solve the high-dimensional problem well, it appears to be powerless in capturing the internal structure of the data. So the subsequent work achieves the purpose of preserving the local geometric structure of the data space by adding graph regularization terms \cite{cai2010graph} as well as popular regularization terms \cite{zong2017multi,wang2018multiview}. To reduce the effect of outliers, $L_{21}$ norm with manifold regularization has to be introduced in work \cite{wu2018manifold}. With the development of research, the information extracted by single-layer NMF clustering often does not meet our needs for data information mining. To explore the deeper and hidden information of data, the approach \cite{trigeorgis_deep_nodate} states a deep semi-NMF model to explore the complex hierarchical information with implicit lower-level hidden attributes. 
Influenced by deep semi-NMF, the model DMVC \cite{zhao2017multi} learns public low-dimensional representations which containing deep information by the instruction of the original data structure. Recently, a multi-view clustering via deep NMF method \cite{huang_auto-weighted_2020} has proposed to learn the optimal weights of each view automatically. Despite the success of existing NMF methods, they can still be improved with the following considerations: $i)$ Fully exploit the raw data for more discriminative information. $ii)$ Focus on both shared and specific information between views. $iii)$ Improved fusion strategy for multi-view information.

In order to address these issues, a novel multi-view clustering method via deep NMF and partition alignment (MVC-DMF-PA) is proposed in this paper. We obtain the base partition matrix after deep semi-NMF firstly while also capturing specific information from different views. Secondly, we maximize the alignment of the consensus partition matrix with a uniformly weighted base partition matrix via an optimal permutation. Finally, we unify the base partition learning and late fusion into a unified framework, hoping to learn a consensus partition matrix for clustering. The main contributions in this paper are summarized as follows:
\begin{itemize}
	\item We propose a deep semi-NMF and partition alignment multi-view clustering approach. In this work, we unify partition learning and late fusion stage into a framework that can mutually facilitate and guide each other to obtain the final common partition matrix for clustering.
	
	\item We decompose the feature matrix to obtain the partition matrix of each view by using a deep semi-NMF framework firstly. Then a late fusion approach is used to learn the fused common partition results by aligning multiple partition matrix finally.
	
	\item The iterative update rules are derived to solve the optimization problem and extensive experiments are conducted on six multi-view datasets. The experimental results show that MVC-DMF-PA has good performance compared with other state-of-the-art methods.
\end{itemize}

The rest structure of this paper is as follows: section \ref{rw} presents the work related to the proposed method. Section \ref{method} details the proposed method and the optimization. Section \ref{exp} shows the experimental results on six public datasets. The conclusions are detailed in the last section. For the convenience of the readers, a summary of the general symbols used in this paper is shown in Table \ref{tab:notions}.
\section{Related Work}\label{rw}

%\subsection{Deep Semi-NMF for representation learning}

Due to the excellent performance in latent feature extraction, NMF and many NMF variants are widely used for clustering. So we start with the brief introduction of NMF and semi-NMF, then introduce deep semi-NMF and the formulation of deep semi-NMF for multi-view clustering.

NMF decomposes the non-negative data matrix $\mathbf{X}_{+}$ into two non-negative matrix $\mathbf{Z}$ and $\mathbf{H}$ of lower rank. NMF can be formulated as follow,
\begin{equation}\label{NMF}
	\min_{\mathbf{Z}, \mathbf{H}} \|\mathbf{X}_{+}-\mathbf{Z}\mathbf{H} \|_F^2,
	\text{s.t.} \mathbf{Z}\geq 0, \mathbf{H}\geq 0,
\end{equation}
We define $\mathbf{X}_{+}\in \mathbb{R}^{d\times n}$ as the non-negative data matrix. where $\mathbf{Z} \in \mathbb{R}^{d\times k}$ can be considered as the cluster centroids and $\mathbf{H} \in \mathbb{R}^{k\times n}$ denotes as cluster indicators. We can find that NMF is inherently related to $K$-means clustering algorithm while keeping the orthogonality constraint \cite{ding2005equivalence}. 
When the input data has mixed signs, we can restrict $\mathbf{H}$ to be non-negative while placing no restriction on the signs of $\mathbf{Z}$. This is called semi-NMF \cite{ding2008convex}:
\begin{equation}\label{semi-NMF}
	\min_{\mathbf{Z},\mathbf{H}} \|\mathbf{X}-\mathbf{Z}\mathbf{H} \|_F^2,
	\text{s.t.} \mathbf{H}\geq 0,
\end{equation}
where $\mathbf{X}\in\mathbb{R}^{d\times n}$ as the data matrix when it has mixed signs. When the goal of semi-NMF is to learn a low-dimensional representation $\mathbf{H} \in \mathbb{R}^{l \times n}$ of the original data matrix, and the range of $l$ is $\left[k, d\right]$. $\mathbf{Z} \in \mathbb{R}^{d\times l}$ can be viewed as a mapping of the original data matrix and the new representation $\mathbf{H}$. In many cases the data we wish to analyze is often rather complex and has a collection of distinct and unknown attributes. So a work \cite{trigeorgis_deep_nodate} proposes the Deep Semi-NMF model, which factorizes a given data matrix $\mathbf{X}$ into $m + 1$ factors, as follow:
\begin{equation}\label{Deep_Semi-NMF}
	\min_{\mathbf{Z}_{i},\mathbf{H}_{i}} \|\mathbf{X}-\mathbf{Z}_{1}\mathbf{Z}_{2}\ldots\mathbf{Z}_{m}\mathbf{H}_{m} \|_F^2,
	\text{s.t.} \mathbf{H}_{i} \geq 0.
\end{equation}
Where $\mathbf{H}_{i}=\mathbf{Z}_{i-1}\mathbf{H}_{i-1}$ ($i>2$). When we use this deep semi-NMF framework for multi-view clustering, we can obtain:
\begin{equation}\label{Multi_Deep_Semi-NMF}
	\min_{\mathbf{Z}_{i},\mathbf{H}_{i}^{(v)}}\sum_{v=1}^{V} \|\mathbf{X}^{(v)}-\mathbf{Z}_{1}^{(v)}\mathbf{Z}_{2}^{(v)}\ldots\mathbf{Z}_{m}^{(v)}\mathbf{H}_{m}^{(v)} \|_F^2,
	{s.t.} \mathbf{H}_{i}^{(v)} \geq 0,
\end{equation}

The meaning of the symbols in Eq. (\ref{Multi_Deep_Semi-NMF}) is shown in Table \ref{tab:notions}. After this many scholars trying to do some research on multi-view clustering based on deep semi-NMF framework. Among them \cite {zhao2017multi} proposed a method which is guided by an intrinsic structure to learn a common representation $\mathbf{H}_{m}$ for clustering. The idea can be formulated as follows,
\begin{equation}\label{MCV_DMF}
	\begin{split}
		\min\limits_{\substack{\mathbf{Z}_{i}^{(v)},\mathbf{H}_{i}^{(v)}\\\mathbf{H}_{m},\mathbf{\alpha}^{(v)}}}\sum_{v=1}^{V}(\alpha^{(v)}&)^\gamma(\|\mathbf{X}^{(v)}-\mathbf{Z}_{1}^{(v)}\mathbf{Z}_{2}^{(v)}\ldots\mathbf{Z}_{m}^{(v)}\mathbf{H}_{m}\|_F^2\\&+\mathbf{\beta}\operatorname{tr}(\mathbf{H}_{m}\mathbf{L}^{(v)}  \mathbf{H}_{m}^{\mathrm{T}})),\\
		\text{s.t.} \mathbf{H}_{i}^{(v)}\geq 0,&\mathbf{H}_{m}\geq 0,\sum_{v=1}^{V}\alpha^{(v)}=1, \alpha^{(v)} \geq 0.
	\end{split}
\end{equation}
Where $\mathbf{H}_{i}^{(v)}=\mathbf{Z}_{i-1}^{(v)}\mathbf{H}_{i-1}^{(v)}$ ($i>2$) and $\mathbf{H}_{m}$ is set as a constraint to enforce multi-view data to share the same representation after multi-layer factorization. $\mathbf{L}^{(v)}$ denotes the Laplacian of the graph for view $v$, which to preserve the geometric structure of origin data. 

Inspired by this work, but we hold a different opinion. We think the representation learned from each view has the unique information of each view. So the new representation cannot be the same exactly, but there must be a same clustering result. In addition, the use of the original structural information will inhibit the learning of the representation to a certain extent and affects the final clustering result. So we propose a novel multi-view clustering algorithm based on late fusion and deep semi-NMF. The details will be introduced in the section \ref{method}.

\section{The Proposed Method}
\label{method}

\begin{table}[t]
	\renewcommand{\arraystretch}{1.3}
	\caption{{Basic notations for the proposed method.}}
	\label{tab:notions} 
	% 	\centering
	\begin{tabular}{ll}
		\toprule
		Notations       & Meaning \\
		\midrule
		$\mathbf{A}$                                                     & A matrix with bold capital symbol\\
		$\mathbf{A}_{i,:}$, $\mathbf{A}_{:,j}$, $\mathbf{A}_{i,j}$     &The $i$-th row, $j$-th column, $ij$-th element\\
		$\mathbf{A}_{m}^{(v)}$  & The $m$-th layer of $v$-th view\\
		$\|\mathbf{A}\|_{F}$, $\operatorname{tr}(\mathbf{A})$
		&The Forbenius norm, trace of matrix $\mathbf{A}$\\
		$\mathbf{A}^{\operatorname{T}}$, $\mathbf{A}^{\dagger}$              
		&The transpose, MP inverse of matrix $\mathbf{A}$\\
		$\left[\mathbf{A}\right]^{+}$, $\left[\mathbf{A}\right]^{-}$    &The negative, positive parts of matrix $\mathbf{A}$\\		
		$\mathbf{X}^{(v)} \in \mathbb{R}^{d_{v}\times n}$   			&Feature matrix of the $v$-th view \\
		$\mathbf{Z}_{1}^{(v)} \in \mathbb{R}^{d_{v}\times l_{1}}$		&$1$-th layer basis matrix of the $v$-th view \\
		$\mathbf{Z}_{i}^{(v)} \in \mathbb{R}^{l_{i-1}\times l_{i}}$	&$i$-th ($i$>1) layer basis matrix of the $v$-th view \\
		$\mathbf{H}_{i}^{(v)} \in \mathbb{R}^{l_{i}\times n}$		& $i$-th layer feature representation of the $v$-th view \\
		$\mathbf{H}_{m}^{(v)} \in \mathbb{R}^{l_{i}\times n}$		& Partition matrix of the $v$-th view\\
		$\mathbf{H} \in \mathbb{R}^{k \times n}$			& Common partition matrix\\
		$\mathbf{W}^{(v)} \in \mathbb{R}^{k\times k}$		&The rotation matrix of the $v$-th view\\
		\bottomrule
	\end{tabular}
\end{table}
In this section, we briefly introduce the motivation of our proposed method firstly. Second, we will talk about our proposed multi-view clustering method based on deep semi-NMF and partition alignment in detail. Finally, we will summarize the overall algorithm and provide an analysis of time complexity. As shown in Table \ref{tab:notions}, we list the symbols used in our work and describe them in addition to the temporary symbols. To make it easier to read, we also explain some necessary symbols in the article.

\subsection{Motivation}

Multi-view clustering is a task of information fusion essentially. As far as we know, information fusion can be divided into early fusion and late-fusion according to the fusion stage, or called feature level and decision level fusion. Although we can get a result at either stage of fusion, the advantage of late fusion is that it reduces the interference of other information channels to every separate partition. So how do we perform late fusion for the base partition we have obtained? The lower right corner of Figure \ref{fig:framework} shows a small demo of the late fusion process. We can find that although $\mathbf{H}_{m}^{(1)}$ and $\mathbf{H}_{m}^{(v)}$ have different representations, both of them has the same the clustering results as they show. We denote $\mathbf{H}$ as the common partition or called consensus partition matrix. The goal of late fusion is to obtain a consensus partition matrix when maximizing the alignment of the consensus partition matrix with a uniformly weighted $W$ base partition matrix via an optimal permutation.

\subsection{Proposed Formulation}
As stated in Section \ref{rw}, we believe that the clustering result from a different view after multi-layer semi-NMF not the representation should be same. So our work is constructed on the basis of deep semi-NMF and late fusion. Unlike the previous work on early fusion, we use late fusion or called decision-level fusion to reduce the effect of random noise. The objective equation of our proposed method is shown below:
\begin{equation}\label{object_all}
	\begin{aligned}
		\min \limits_{\substack{\mathbf{Z}_{i}^{(v)},\mathbf{H}_{i}^{(v)},\mathbf{H}\\\mathbf{\alpha}^{(v)},\mathbf{\beta}^{(v)},\mathbf{W}^{(v)}}}
		&\sum_{v=1}^{V}(\alpha^{(v)})^2\|\mathbf{X}^{(v)}-\mathbf{Z}_{1}^{(v)}\mathbf{Z}_{2}^{(v)}\ldots\mathbf{Z}_{m}^{(v)}\mathbf{H}_{m}^{(v)}\|_F^2-\\
		&\lambda\operatorname{Tr}(\mathbf{H}\sum_{v=1}^{V}\beta^{(v)}\mathbf{H}_{m}^{(v)\mathrm{T}}{\mathbf{W}^{(v)}}), \\
		\text{s.t.} \mathbf{H}_{i}^{(v)} &\geq 0,\mathbf{H}\mathbf{H}^{\operatorname{T}}=\mathbf{I}_{k},\mathbf{W}^{(v)}\mathbf{W}^{(v)\operatorname{T}}=\mathbf{I}_{k},\sum_{v=1}^{V}\alpha^{(v)}=1, \\
		& \alpha^{(v)} \geq 0,{\vert| \beta |\vert}_2=1, \beta^{(v)} \geq 0.
	\end{aligned}
\end{equation}
The first term of the optimization objective represents the reconstruction loss of $V$ views, which is the objective equation of the multi-view deep semi-NMF. $\alpha$ represents the percentage of reconstruction loss for all views. The dimension of the last layer is $k$, which means that $\mathbf{H}_{m}^{(v)}$ represents the partition matrix of the $v$-th view. To accommodate each dataset, we adjust the dimensions of the different layers as multiples of the number of clusters. This reconstruction term loss can help us to explore more rich hidden information of origin data better. Different views have different origins, so there will be some differences in the final partition matrix for each view as explained in Section \ref{rw} and we denote as $\mathbf{H}_{m}^{(v)}$. 

The second term of optimization objective represents the loss of late fusion. $\mathbf{H}_{m}^{(v)}$ represents the partition matrix of the $v$-th view and $\mathbf{H}$ denotes the consensus clustering partition matrix. $\mathbf{W}^{(v)}$ denotes the column alignment matrix of the $v$-th view and this matrix can do the column exchanges to solve the case that the clustering index matrix of different views have the same meaning but different representation. $\beta^{(v)}$ is the weighting coefficient of the $v$-th partition matrix for fusing into $\mathbf{H}$. Therefore, the objective function of late fusion is to maximize the alignment of the consensus partition matrix $\mathbf{H}$ and fused partition matrix $\sum_{v=1}^{V}\beta^{(v)}\mathbf{H}_{m}^{(v)\mathrm{T}}{\mathbf{W}^{(v)}}$.

\subsection{Initialization}\label{init}
Following the initialization of $\mathbf{Z}_{i}^{(v)}$ and $\mathbf{H}_{i}^{(v)}$ of the work \cite{zhao2017multi}, we also do it layer by layer. First, we decompose the data matrix $\mathbf{X}^{(v)}$ of view$v$ : $\mathbf{X}^{(v)}\approx\mathbf{Z}_{1}^{(v)}\mathbf{H}_{1}^{(v)}$ to get the new representation $\mathbf{H}_{1}^{(v)}$. Then we decompose $\mathbf{H}_{1}^{(v)}$: $\mathbf{H}_{1}^{(v)}\approx\mathbf {Z}_{2}^{(v)}\mathbf{H}_{2}^{(v)}$ to get $\mathbf{H}_{2}^{(v)}$. Continue to decompose the new representation obtained until we get the partition matrix $\mathbf{H}_{m}^{(v)}$. Finally, we get $\mathbf{H}_{m}^{(v)}$($v=1\dots V$) for all views. By setting $\mathbf{W}^{(v)} = \mathbf{I}_k$, we get the initialization of $\mathbf{W}$ and the condition $\mathbf{W}^{(v)}\mathbf{W}^{(v)\operatorname{T}}=\mathbf{I}_{k}$ is satisfied. We consider the losses contributed by all views to be the same at the beginning, so we set $\alpha^{(v)} = {1}/{V}$ and $\beta^{(v)} = {1}/{\sqrt{V}}$. 

\subsection{Optimization}
In order to solve Eq. \ref{object_all}, we designed a seven-step alternate optimization algorithm, where three steps are inherited from the optimization of the original depth semi-NMF, two steps can be easily solved by off-the-shelf packages, and the last two steps can be derived as closed-form solutions. The point to note is that for the $v$-th view, we need to optimize $\mathbf {Z}_{i}^{(v)}$ and $\mathbf{H}_{i}^{(v)}$ layer by layer, i.e., first $\mathbf {Z}_{1}^{(v)}$ and then $\mathbf{H}_{1}^{(v)}$ until $\mathbf{Z}_{m}^{(v)}$ and $\mathbf{H}_{m}^{(v)}$ 
are optimized.

%%%%%---------优化H-------------%%%%%%%%	
\subsubsection{Subproblem of updating $\mathbf{H}$}
\ 
\newline
With $\mathbf{Z}_{i}^{(v)}$, $\mathbf{H}_{i}^{(v)}$, $\mathbf{W}^{(v)}$, $\alpha$ and $\beta$ fixed, the optimization Eq. (\ref{object_all}) can be written as follow,
\begin{equation}\label{upH}
	\mathcal{C}=-\operatorname{Tr}(\mathbf{H}\mathbf{U}),
	\text{s.t.} \mathbf{H}\mathbf{H}^{\operatorname{T}}=\mathbf{I}_{k}.
\end{equation}	
Where $\mathbf{U}$=$\sum_{v=1}^{V}\beta^{(v)}\mathbf{H}_{m}^{(v)\mathrm{T}}{\mathbf{W}^{(v)}}$. This problem in Eq. (\ref{upH})
could be easily solved by taking the singular value decomposition (SVD) of the given matrix U.
% 

%%%%%---------优化Zi--------------%%%%%%%%	
\subsubsection{Subproblem of updating $\mathbf{Z_{i}^{(v)}}$}
\ 
\newline
With $\mathbf{H}$, $\mathbf{H}_{i}^{(v)}$, $\mathbf{W}^{(v)}$, $\alpha$ and $\beta$ fixed, the optimization Eq. (\ref{object_all}) can be written as follow,
\begin{equation}\label{obj_z}
	\begin{array}{l}
		\mathcal{C}=\|\mathbf{X}^{(v)}-\phi\mathbf{Z}_{i}^{(v)}\mathbf{H}_{i}^{(v)}\|_F^2,\\
	\end{array}
\end{equation}
where $\phi=\mathbf{Z}_{1}^{(v)}\mathbf{Z}_{2}^{(v)}\ldots\mathbf{Z}_{i-1}^{(v)}$, by setting $\partial\mathcal{C}/\partial \mathbf{Z}_{i}^{(v)}=0$, we can easily obtain the solutions as:
\begin{equation}\label{upZ}
	\mathbf{Z}_{i}^{(v)}=\phi^{\dagger} \mathbf{X}^{(v)} \mathbf{H}_{i}^{(v) \dagger}.
\end{equation}	

%%%%%-------------------优化Hi-------------------%%%%%%%%	
\subsubsection{Subproblem of updating $\mathbf{H}_{i}^{(v)}(i \textless m)$}	
\ 
\newline
With $\mathbf{H}$, $\mathbf{Z}_{i}^{(v)}$, $\mathbf{W}^{(v)}$, $\alpha$ and $\beta$ fixed, the optimization Eq. (\ref{object_all}) can be written as follows,

\begin{equation}\label{Hi}
	\mathcal{C}=\|\mathbf{X}^{(v)}-\Phi\mathbf{H}_{i}^{(v)}\|_F^2,\text{s.t.} \mathbf{H}_{i}^{(v)} \geq 0.
\end{equation}
where $\Phi=\mathbf{Z}_{1}^{(v)}\mathbf{Z}_{2}^{(v)}\ldots\mathbf{Z}_{i}^{(v)}$, Following \cite{ding_convex_2010}, the update rule for $\mathbf{H}_{i}^{(v)}(i \textless m)$ is
\begin{equation}\label{upHi}
	\mathbf{H}_{i}^{(v)}=\mathbf{H}_{i}^{(v)} \odot \sqrt{{\left[\Phi^{\mathrm{T}} \mathbf{X}^{(v)}\right]^{\mathrm{+}}+\left[\Phi^{\mathrm{T}} \Phi \mathbf{H}_{i}^{(v)}\right]^{\mathrm{-}}}/{\left[\Phi^{\mathrm{T}} \mathbf{X}^{(v)}\right]^{\mathrm{-}}+\left[\Phi^{\mathrm{T}} \Phi \mathbf{H}_{i}^{(v)}\right]^{\mathrm{+}}}}.
\end{equation}

where $\left[\mathbf{A}\right]^{\mathrm{+}}=(\vert\mathbf{A}\vert+\mathbf{A})/2$, $\left[\mathbf{A}\right]^{\mathrm{-}}=(\vert\mathbf{A}\vert-\mathbf{A})/2$. Same to our previous work \cite{zhao2017multi}, we usually update $\mathbf{H}_{m}^{(v)}$ with the above update rule before using the update rule of $\mathbf{H}_{i}^{(v)}$ in order to facilitate the code writing and to allow the program convergence faster.

%%%%%------------优化Hm-----------%%%%%%%	

\subsubsection{Subproblem of updating $\mathbf{H}_{m}^{(v)}$ }	
\ 
\newline
With $\mathbf{H}$, $\mathbf{Z}_{i}^{(v)}$, $\mathbf{H}_{i}^{(v)}$ ($i \textless m$), $\mathbf{W}^{(v)}$, $\alpha$ and $\beta$ fixed, the optimization Eq. (\ref{object_all}) can be written as follow,
\begin{equation}\label{reHm}
	\mathcal{C}=\|\mathbf{X}^{(v)}-\Phi\mathbf{H}_{m}^{(v)}\|_F^2
	-\lambda\operatorname{Tr}(\beta^{(v)}\mathbf{H}\mathbf{H}_{m}^{(v)\mathrm{T}}\mathbf{W}^{(v)}+\mathbf{G})\quad\text{s.t.} \mathbf{H}_{m}^{(v)} \geq 0,
\end{equation}
where $\Phi$=$\mathbf{Z}_{1}^{(v)}\mathbf{Z}_{2}^{(v)}\ldots\mathbf{Z}_{m}^{(v)}$ and $\mathbf{G}$=$\sum_{o=1,o\neq v}^{V}\beta^{(o)}\mathbf{H}\mathbf{H}_{m}^{(o)\mathrm{T}}\mathbf{W}^{(o)}$, We give the updating rule of $\mathbf{H}_{m}^{(v)}$ first, followed by the proof of it.
\begin{equation}\label{upHm}
	\begin{split}
		\mathbf{H}_{m}^{(v)}=&\mathbf{H}_{m}^{(v)} \odot \sqrt{
			{\vartheta_{u}(\mathbf{ZHW})}/
			{\vartheta_{u}(\mathbf{ZHW})}},\\
		\vartheta_{u}(\mathbf{ZHW})=&2(\alpha^{(v)})^{2}(\left[\Phi^{\mathrm{T}} \mathbf{X}^{(v)}\right]^{\mathrm{+}}+\left[\Phi^{\mathrm{T}}\Phi\mathbf{H}_{m}^{(v)}\right]^{\mathrm{-}})+\lambda\beta^{(v)}\left[\mathbf{W}^{(v)}\mathbf{H}\right]^{\mathrm{+}},\\
		\vartheta_{u}(\mathbf{ZHW})=&2(\alpha^{(v)})^{2}(\left[\Phi^{\mathrm{T}} \mathbf{X}^{(v)}\right]^{\mathrm{-}}+\left[\Phi^{\mathrm{T}}\Phi\mathbf{H}_{m}^{(v)}\right]^{\mathrm{+}})+\lambda\beta^{(v)}\left[\mathbf{W}^{(v)}\mathbf{H}\right]^{\mathrm{-}}.
	\end{split}
\end{equation}

Next we prove that Eq. (\ref{upHm}) is a solution of Eq. (\ref{reHm}). We introduce the Lagrangian function of Eq. (\ref{reHm}) as follow,
\begin{align*}
	\mathbf{L}(\mathbf{H}_{m}^{(v)})=&(\alpha^{(v)})^{2}\|\mathbf{X}^{(v)}-\Phi\mathbf{H}_{m}^{(v)}\|_F^2	-\lambda\operatorname{Tr}(\beta^{(v)}\mathbf{H}\mathbf{H}_{m}^{(v)\mathrm{T}}\mathbf{W}^{(v)}+\mathbf{G})-\\
	&\operatorname{Tr}(\eta\mathbf{H}_{m}^{(v)}).
\end{align*}
By setting $\partial\mathbf{L}(\mathbf{H}_{m}^{(v)})/\partial \mathbf{H}_{m}^{(v)}=0$, From the complementary slackness condition, we can obtain
\begin{align*}
	&(-2(\alpha^{(v)})^{2}\Phi^\mathrm{T} \mathbf{X}^{(v)}+2(\alpha^{(v)})^{2} \Phi^\mathrm{T} \Phi\mathbf{H}_{m}^{(v)}\\&-\lambda\beta^{(v)}\mathbf{W}^{(v)}\mathbf{H})\mathbf{H}_{m}^{(v)}=\eta\mathbf{H}_{m}^{(v)}=0.
\end{align*}
So we can obtain:
\begin{align*}
	&(-2(\alpha^{(v)})^{2}\Phi^\mathrm{T} \mathbf{X}^{(v)}+2(\alpha^{(v)})^{2} \Phi^\mathrm{T} \Phi\mathbf{H}_{m}^{(v)}-\lambda\beta^{(v)}\mathbf{W}^{(v)}\mathbf{H})\mathbf{H}_{m}^{(v)2}=0.
\end{align*}
Then we can easily get the update rule Eq. (\ref{reHm} ) for $\mathbf{H}_{m}^{(v)}$.

%%%%%------------------优化W---------------%%%%%%%%	
\subsubsection{Subproblem of updating $\mathbf{W}^{(v)}$}	
\ 
\newline
With $\mathbf{H}$, $\mathbf{Z}_{i}^{(v)}$, $\mathbf{H}_{i}^{(v)}$, $\alpha$ and $\beta$ fixed, the optimization Eq. \ref{object_all} can be written as follows,
\begin{equation}\label{upW}
	\mathcal{C}=-\operatorname{Tr}(\mathbf{W}^{(v)\operatorname{T}}\mathbf{Q}), \quad
	\text{s.t.}\mathbf{W}^{(v)}\mathbf{W}^{(v)\operatorname{T}}=\mathbf{I}_{k}.
\end{equation}
Where $\mathbf{Q}=\beta^{(v)}\mathbf{H}_{m}^{(v)}\mathbf{H}^{\operatorname{T}}$. This problem in Eq.(\ref{upW}) could be easily solved by taking the singular
value decomposition (SVD) of the given matrix $\mathbf{Q}$.

%%%%%----------------优化α---------------%%%%%%%%	
\subsubsection{Subproblem of coefficient updating $\alpha^{(v)}$}
\ 
\newline
With $\mathbf{H}$, $\mathbf{Z}_{i}^{(v)}$, $\mathbf{H}_{i}^{(v)}$, $\mathbf{W}^{(v)}$ and $\beta$ fixed, the optimization Eq. \ref{object_all} can be written as following,
\begin{equation}\label{upalpha}
	\mathcal{C}=(\alpha^{(v)})^2\mathcal{R}^{(v)},\quad\text{s.t.} \sum_{v=1}^{V}\alpha^{(v)}=1, \alpha^{(v)} \geq 0.
\end{equation}
Suppose that $\mathcal{R}^{(v)}=\|\mathbf{X}^{(v)}-\mathbf{Z}_{1}^{(v)}\mathbf{Z}_{2}^{(v)}\ldots\mathbf{Z}_{m}^{(v)}\mathbf{H}_{m}^{(v)}\|_F^2$. The Lagrange function of Eq. (\ref{upalpha}) can be rewritten as:
\begin{equation}\label{Lalpha}
	\mathbf{L}(\alpha^{(v)})=(\alpha^{(v)})^2\mathcal{R}^{(v)}-\gamma(\sum_{v=1}^{V}\alpha^{(v)}-1).
\end{equation}
Where $\gamma$ is the Lagrange multiplier. By taking the derivative of Eq. (\ref{Lalpha}) with respect to $\alpha^{(v)}$ and setting it to zero, we can obtain $\alpha^{(v)}={\gamma}/{2\mathcal{R}^{(v)}}$. Then we replace $\alpha^{(v)}$ in Eq. (\ref{upalpha}) into $\sum_{v=1}^{V}\alpha^{(v)}=1$ and finally obtain $\alpha^{v}$ as follows,
\begin{equation}\label{upalpha1}
	\alpha^{(v)}={\sum_{v=1}^{V}\mathcal{R}^{(v)}}/{\mathcal{R}^{(v)}}.
\end{equation}

%%%%%-----------------优化β--------------------%%%%%%%%	
\subsubsection{Subproblem of updating coefficient $\beta$}
\ 
\newline
With $\mathbf{H}$, $\mathbf{Z}_{i}^{(v)}$, $\mathbf{H}_{i}^{(v)}$, $\mathbf{W}^{(v)}$ and $\alpha$ fixed, the optimization Eq. \ref{object_all} can be written as following,
\begin{equation}\label{upbeta}
	\begin{array}{l}
		\max \operatorname{Tr}(\sum_{v=1}^{V}\beta^{(v)}\mathbf{H}_{m}^{(v)\mathrm{T}}{\mathbf{W}^{(v)}}\mathbf{H}) \text{s.t.} {\vert| \beta |\vert}_2=1, \beta^{(v)} \geq 0.
	\end{array}
\end{equation}
The objection can be written as follow,
\begin{equation}\label{re_beta}
	\begin{array}{l}
		\max \limits_{\beta}\mathbf{f}^\mathrm{T}\beta, \quad
		\text{s.t.} {\vert| \beta |\vert}_2=1, \beta \geq 0,
	\end{array}       
\end{equation}    
where $\mathbf{f}^{\mathrm{T}}$=$\left[f_{1}, f_{2}, \ldots, f_{V}\right]$ with $\mathbf{f}_{v}$=$\operatorname{Tr}(\mathbf{H}_{m}^{(v)T}{\mathbf{W}^{(v)}}\mathbf{H})$. According to Cauchy-inequality, the update rule of $\beta$ as follow,
\begin{equation}
	\beta = \mathbf{f}/\sqrt{\sum{\mathbf{f}^{2}}}.
\end{equation}
We summarize the proposed algorithm in Algorithm \ref{Algorithm-proposed}. We train the proposed algorithm at least 150 iterations until convergence, then we perform $K$-means on $\mathbf{H}$ to obtain the clustering results.
\alglanguage{pseudocode}
\begin{algorithm}[t]
	\caption{MVC-DMF-PA}
	\label{Algorithm-proposed}
	\begin{algorithmic}[1]
		\Require 
		$\{\mathbf{X}^{(v)}\}_{v=1}^{V}$: set of given multi-view data matrices;
		$\lambda$: the parameter for balancing reconstruction loss and late fusion loss;
		$p$: parameters of the corresponding dimensions for different layers;
		\State Initialize $\mathbf{H}_{i}^{(v)}$, $\mathbf{Z}_{i}^{(v)}$, $\mathbf{W^{(v)}}$, $\alpha$ and $\beta$ according to section \ref{init}.
		\While{not\;convergence}
		\State update $\mathbf{H}$ by solving Eq. (\ref{upH}).
		\For{v $\leq$ V}
		\For{i $\leq$ m}
		\State update $\mathbf{Z}_{i}^{(v)}$ by solving Eq. (\ref{upZ}).
		\State update $\mathbf{H}_{i}^{(v)}$ by solving Eq. (\ref{upHi}).
		\EndFor
		\State update $\mathbf{H}_{m}^{(v)}$ by solving Eq. (\ref{upHm}).
		\EndFor
		\For{v $\leq$ V}
		\State update $\mathbf{W}^{(v)}$ by solving Eq. (\ref{upW}).
		\EndFor
		\For{v $\leq$ V}
		\State update $\alpha^{(v)}$ by solving Eq. (\ref{upalpha}).
		\EndFor
		\State update $\beta$ by solving Eq. (\ref{upbeta}).
		\EndWhile 
		\State \Return Consensus partition matrix $\mathbf{H}.$ Performing $K$-means on $\mathbf{H}$ to get final clustering result.
	\end{algorithmic}
\end{algorithm}

\begin{table*}[t]
	\caption{ACC, NMI and purity comparison of different clustering algorithms on all datasets. The best results are highlighted in bold}
	\label{tab:res_all} 
	\centering
	\resizebox{\textwidth}{45mm}{
		\begin{tabular}{cccccccccccccc}
			\toprule
			Datasets & CKM    & Co-train & Co-reg & MVKKM  & MultiNMF & DMVC   & MVCF   & ScaMVC & GMC    & AwDMVC & CSMVSC & PMSC   & OURS            \\
			\midrule	
			\multicolumn{14}{c}{ACC} \\
			\midrule
			BBC               & 0.4036 & 0.3271   & 0.4061 & 0.4492 & 0.4826   & 0.4948 & 0.6575 & 0.5195 & 0.6934 & 0.6504 & 0.4745 & 0.3664 & \textbf{0.8102} \\
			BBCSport          & 0.4797 & 0.3918   & 0.2962 & 0.4045 & 0.5751   & 0.4381 & 0.6324 & 0.4367 & 0.7390 & 0.7076 & 0.4651 & 0.3750 & \textbf{0.9375} \\
			MSRCV1            & 0.3238 & 0.8114   & 0.8110 & 0.6905 & -        & 0.4048 & 0.8952 & 0.4190 & 0.8952 & -      & 0.3524 & 0.3238 & \textbf{0.9143} \\
			ORL               & 0.5825 & 0.7250   & 0.8325 & 0.6250 & 0.2375   & 0.7700 & 0.6650 & 0.6175 & 0.6325 & 0.1200 & 0.2275 & 0.1850 & \textbf{0.8675} \\
			Reuters           & 0.3900 & 0.5268   & 0.4699 & 0.2208 & 0.3633   & 0.3233 & 0.1675 & 0.1625 & 0.1992 & 0.3408 & 0.2575 & 0.1692 & \textbf{0.5908} \\
			HW                & 0.6490 & 0.8015   & 0.8204 & 0.6190 & 0.7854   & 0.3870 & 0.1005 & 0.7520 & 0.7610 & 0.2875 & 0.8065 & 0.6515 & \textbf{0.8690} \\
			Average Rank      & 8.17   & 6.33     & 5.83   & 8.50   & 7.33     & 7.17   & 6.50   & 8.17   & 4.67   & 8.33   & 7.67   & 10.83  & \textbf{1.00}  \\
			\midrule
			\multicolumn{14}{c}{NMI} \\
			\midrule
			BBC          & 0.2206          & 0.1094   & 0.1128 & 0.2096 & 0.2737   & 0.2016 & 0.4280 & 0.2018 & 0.4852          & 0.4574 & 0.1828 & 0.0555 & \textbf{0.6406} \\
			BBCSport     & 0.2764          & 0.1648   & 0.1318 & 0.1909 & 0.3796   & 0.2604 & 0.4045 & 0.2036 & 0.7047          & 0.4682 & 0.1224 & 0.0278 & \textbf{0.8178} \\
			MSRCV1       & 0.7564          & 0.7434   & 0.7293 & 0.5672 & -        & 0.2200 & 0.8137 & 0.6537 & 0.8189          & -      & 0.1898 & 0.2681 & \textbf{0.8536} \\
			ORL          & 0.7722          & 0.8661   & 0.9106 & 0.7797 & 0.3798   & 0.8800 & 0.8102 & 0.7892 & 0.8035          & 0.4343 & 0.3837 & 0.3553 & \textbf{0.9284} \\
			Reuters      & \textbf{0.3942} & 0.3129   & 0.2720 & 0.1035 & 0.3220   & 0.1348 & 0.0306 & 0.0306 & 0.0820          & 0.3056 & 0.0803 & 0.0042 & 0.3715 \\
			HW           & 0.6223          & 0.7659   & 0.7626 & 0.6564 & 0.7464   & 0.3865 & 0.0045 & 0.7564 & \textbf{0.8118} & 0.6293 & 0.7568 & 0.6165 & 0.7658 \\
			Average Rank & 6.00            & 6.17     & 6.67   & 8.00   & 7.33     & 8.00   & 6.67   & 7.83   & 3.67            & 7.00   & 9.83   & 12.00  & \textbf{1.50}\\
			\midrule
			\multicolumn{14}{c}{PUR}\\
			\midrule
			BBC          & 0.4063          & 0.3315   & 0.3424 & 0.4635 & 0.4825   & 0.4838 & 0.6584 & 0.5256 & 0.6934          & 0.7755 & 0.4876 & 0.3693 & \textbf{0.8102} \\
			BBCSport     & 0.4936          & 0.4368   & 0.3631 & 0.3761 & 0.5923   & 0.5136 & 0.6342 & 0.4426 & 0.7629          & 0.6599 & 0.4779 & 0.3805 & \textbf{0.9375} \\
			MSRCV1       & 0.8524          & 0.8271   & 0.8238 & 0.6905 & -        & 0.4190 & 0.8952 & 0.7429 & 0.8952          & -      & 0.3619 & 0.3333 & \textbf{0.9143} \\
			ORL          & 0.6300          & 0.7668   & 0.8500 & 0.6850 & 0.2375   & 0.7975 & 0.6850 & 0.6600 & 0.7150          & 0.1200 & 0.2975 & 0.2400 & \textbf{0.8875} \\
			Reuters      & 0.5458 & 0.5378   & 0.4816 & 0.2633 & 0.4533   & 0.3358 & 0.1708 & 0.1708 & 0.2417          & 0.4875 & 0.2675 & 0.1708 & \textbf{0.5908} \\
			HW           & 0.6830          & 0.8092   & 0.8258 & 0.6550 & 0.7981   & 0.3860 & 0.2000 & 0.7520 & 0.7825 & 0.5345 & 0.8175 & 0.6625 & \textbf{0.8690} \\
			Average Rank & 6.67            & 6.50     & 6.67   & 9.00   & 8.00     & 7.33   & 6.67   & 7.83   & 4.67            & 7.50   & 7.50   & 10.67  & \textbf{1.00}  \\
			\bottomrule
	\end{tabular}}
\end{table*}

\begin{table}[]
	\caption{Datasets used in our experiments.}
	\label{tab:datasets} 
	\resizebox{0.49\textwidth}{!}{
		\begin{tabular}{cccccc}
			\toprule
			Dataset  & Type & Views number & View Dimension  & Sample number & Cluster number \\
			\midrule
			BBC      & text          & 4                     & 4659 4633 4665 4684      & 685                    & 5                       \\
			BBCSport & text & 2            &3183 3203      & 544           & 5              \\
			MSRCV1    & image         & 5                     & 1302 512 100 256 210     & 210                    & 7                       \\
			ORL      & image         & 3                     & 4096 3304 6750           & 400                    & 40                      \\
			Reuters  & text          & 5                     & 2000 2000 2000 2000 2000 & 1200                   & 6                       \\
			HW       & image         & 2                     & 240 216                  & 2000                   & 10 						\\
			\bottomrule                      
	\end{tabular}}%}
\end{table}

\subsection{Computational Complexity}
Our work includes the process of pre-training and fine-tuning, so we will analyze them separately. To make the analysis clearer, we assume the dimensions in all the layers are the same. So we denote $l$. The dimensions of the original feature for all the views are the same which denoted $d$. $t_{pre}$ denotes the number of iterations to achieve convergence in pre-training process and $t_{fine}$ denotes the number of iterations to achieve convergence in fine-tuning process. So the complexity of pre-training and fine-tuning stages are $O(Vmt_{pre}(nd^2 + dnl + ldn + lp^{2} + lpn))$ and $O(Vmt_{fine}(ldn+dl^2+nl^2+nk^2+k^3+kn^2))$ respectively, where $l \le d$ and $k<n$ normally. In conclusion, the time complexity of our algorithm is $O(Vmt_{pre}( dl^2 ++ nd^2)$) + $O(Vmt_{fine}(ldn + dl^2 + nl^2 + kn^2))$.

\section{Experiments} \label{exp}
In this section, we present the benchmark dataset and comparison algorithm used for the experiments first, followed by the evaluation of the experimental results, analysis of parameter sensitivity and convergence of our proposed method.
\subsection{Dataset}
We evaluate the performance of the proposed method on six widely-adapted multi-view learning benchmark datasets. There are three image datasets include MSRCV1, YALE, HW and three text datasets BBC, BBCSport, Reuter. The details of these datasets are shown in Table \ref{tab:datasets}.

\subsection{Compared method}
Several representative models are compared in our experiment, including a baseline with all view be concreted $K$-means $\textbf{CKM}$, a kernel-based method $\textbf{MVKKM}$ \cite{tzortzis2012kernel}, a graph-based method $\textbf{GMC}$ \cite{wang2019gmc}, two subspace-based $\textbf{PMSC}$ \cite{kang2020partition} and $\textbf{CSMVSC}$ \cite{luo2018consistent}, two co-training methods $\textbf{Co-train}$ \cite{kumar_co-training_nodate} and $\textbf{Co-reg}$ \cite{kumar_co-regularized_nodate}, and five matrix decomposition representative models $\textbf{MultiNMF}$ \cite{wang2015feature}, $\textbf{MVCF}$ \cite{zhan2018adaptive}, $\textbf{ScaMVC}$ \cite{huang2018self}, $\textbf{DMVC}$ \cite{zhao2017multi} and $\textbf{AwDMVC}$ \cite{huang_auto-weighted_2020}.
\subsection{Experimental setup}
For the proposed method and all compared methods, we perform data pre-processing first, i.e., we normalize all datasets. We consider the number of clusters $k$ as the true number of classes per dataset. For the method we proposed, The weighting coefficient $\gamma$ is selected from $\left[2^{-12}, 2^{-11}, \dots, 2^{4}, 2^{5} \right]$. We assume that the layer size should be correlated with the number of clusters, so we designed two schemes. One layer size $p_2 = \left[l_1, k\right]$ and another layer size $p_3 = \left[l_1, l_2, k\right]$.  Where $l_1$ in $p2$ is chosen from $[4k, 5k, 6k]$ and $l_1$, $l_2$ in $p_3$ are chosen from $[8k, 10k, 12k]$ and $[4k, 5k, 6k]$ respectively. As for these compared methods, we obtain their paper and code from the author's websites and obey the setting of the hyper-parameters in the paper. Three popular metrics are applied to evaluate the clustering performance. They are accuracy (ACC), normalized mutual information (NMI), and purity (PUR). We repeat each experiment 50 times to avoid the effect of the random initialization and save the best result. All experiments are conducted on a desktop computer with Intel i9-9900K CPU@ 3.60GHz×16 and 64GB RAM, MATLAB 2018a (64bit).

\begin{figure*}
	\centering
	\begin{minipage}{0.32\linewidth}
		\vspace{3pt}
		\centerline{\includegraphics[width=\textwidth]{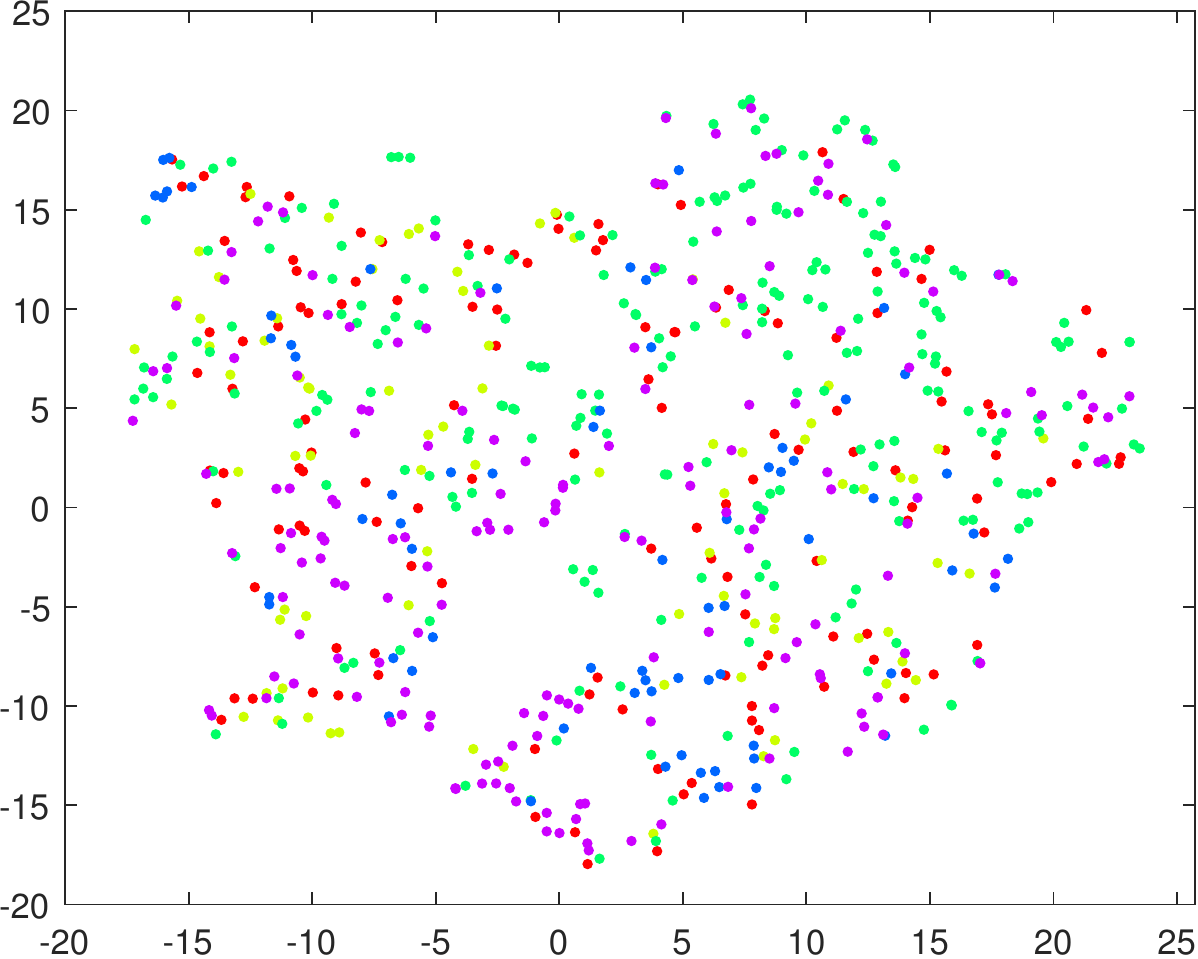}}
		\centerline{(a1) BBC(iter=1)}
	\end{minipage}
	\begin{minipage}{0.32\linewidth}
		\vspace{3pt}
		\centerline{\includegraphics[width=\textwidth]{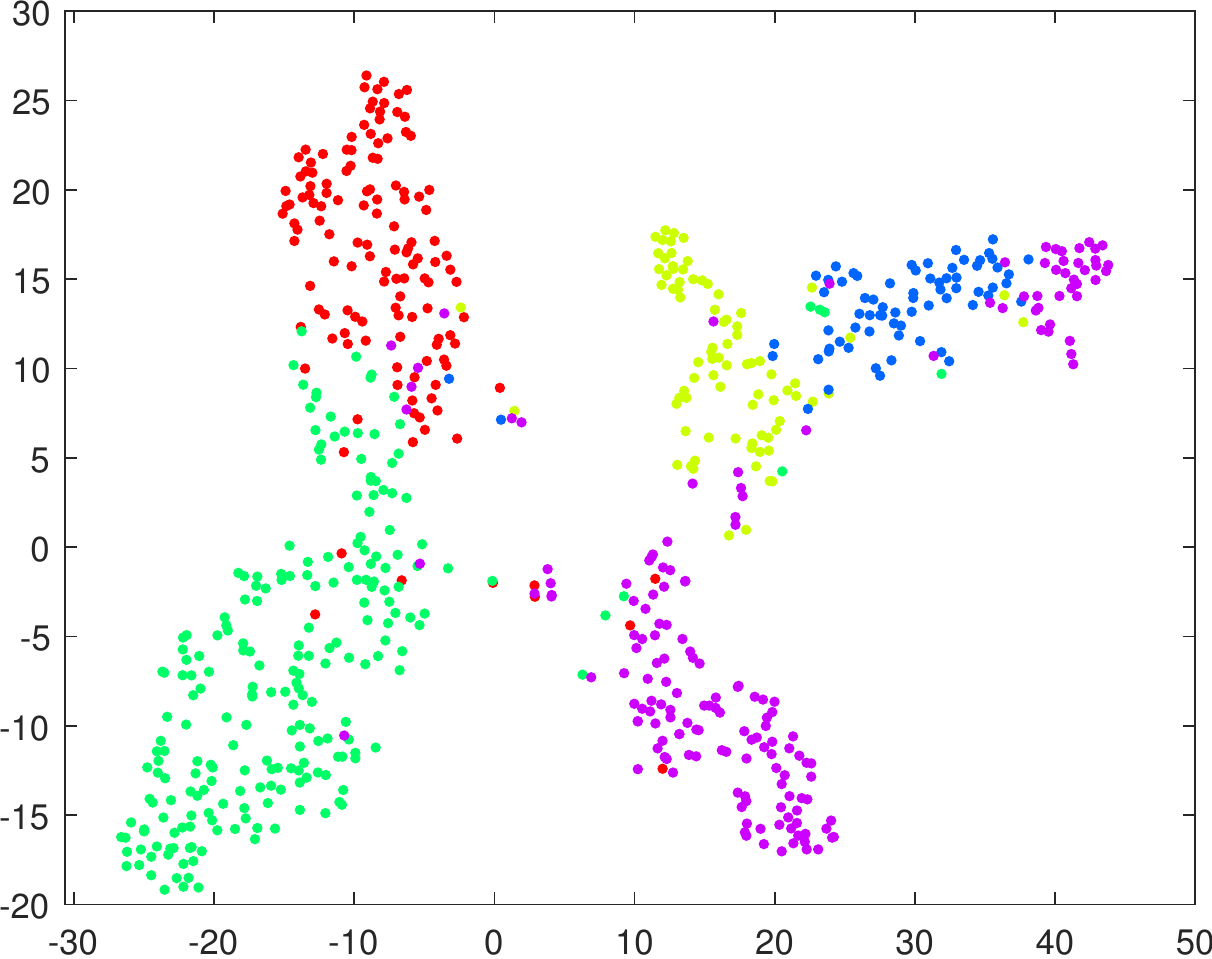}}
		\centerline{(a2) BBC(iter=10)}
	\end{minipage}
	\begin{minipage}{0.32\linewidth}
		\vspace{3pt}
		\centerline{\includegraphics[width=\textwidth]{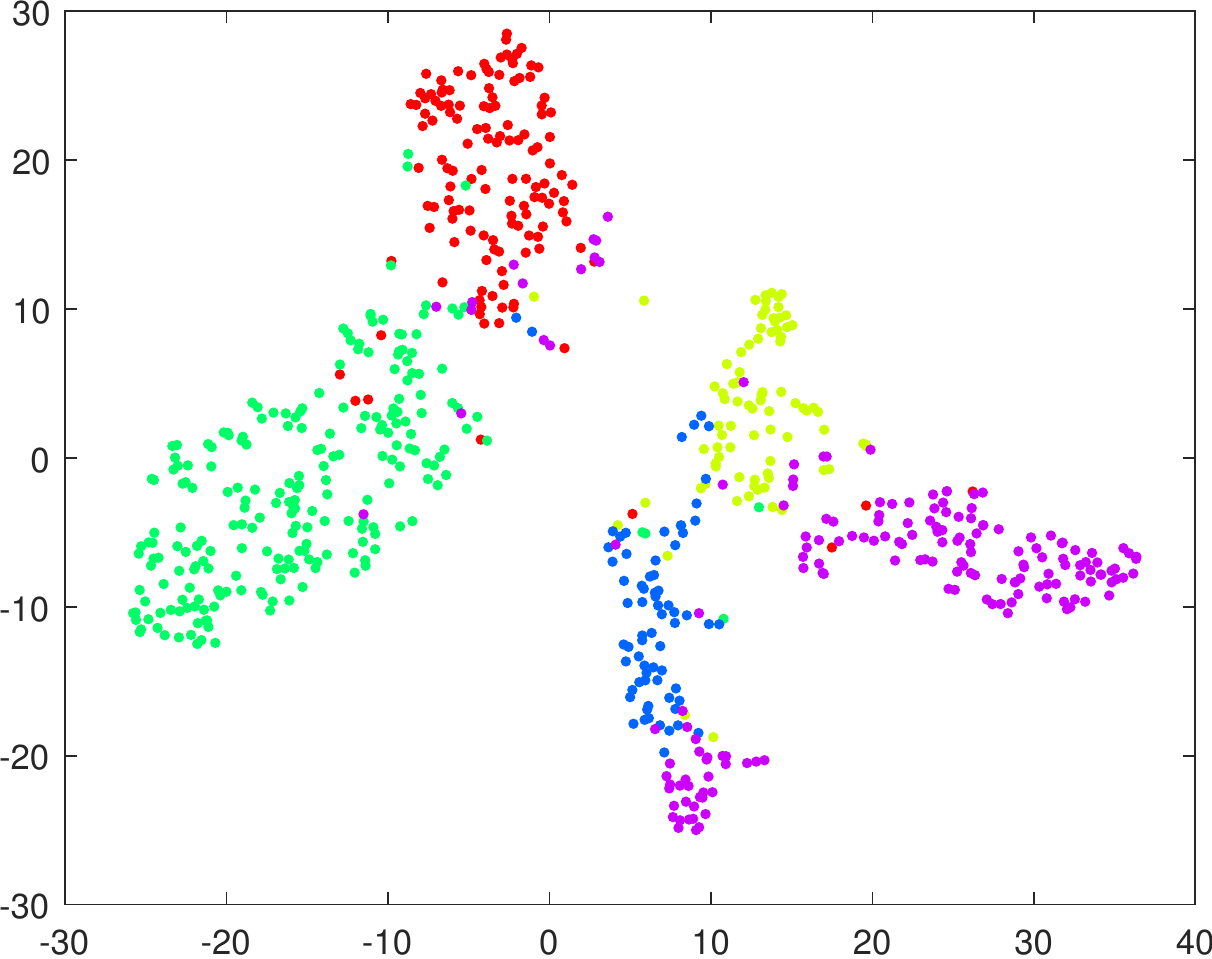}}
		\centerline{(a3) BBC(iter=60)}
	\end{minipage}\\
	\begin{minipage}{0.32\linewidth}
		\vspace{3pt}
		\centerline{\includegraphics[width=\textwidth]{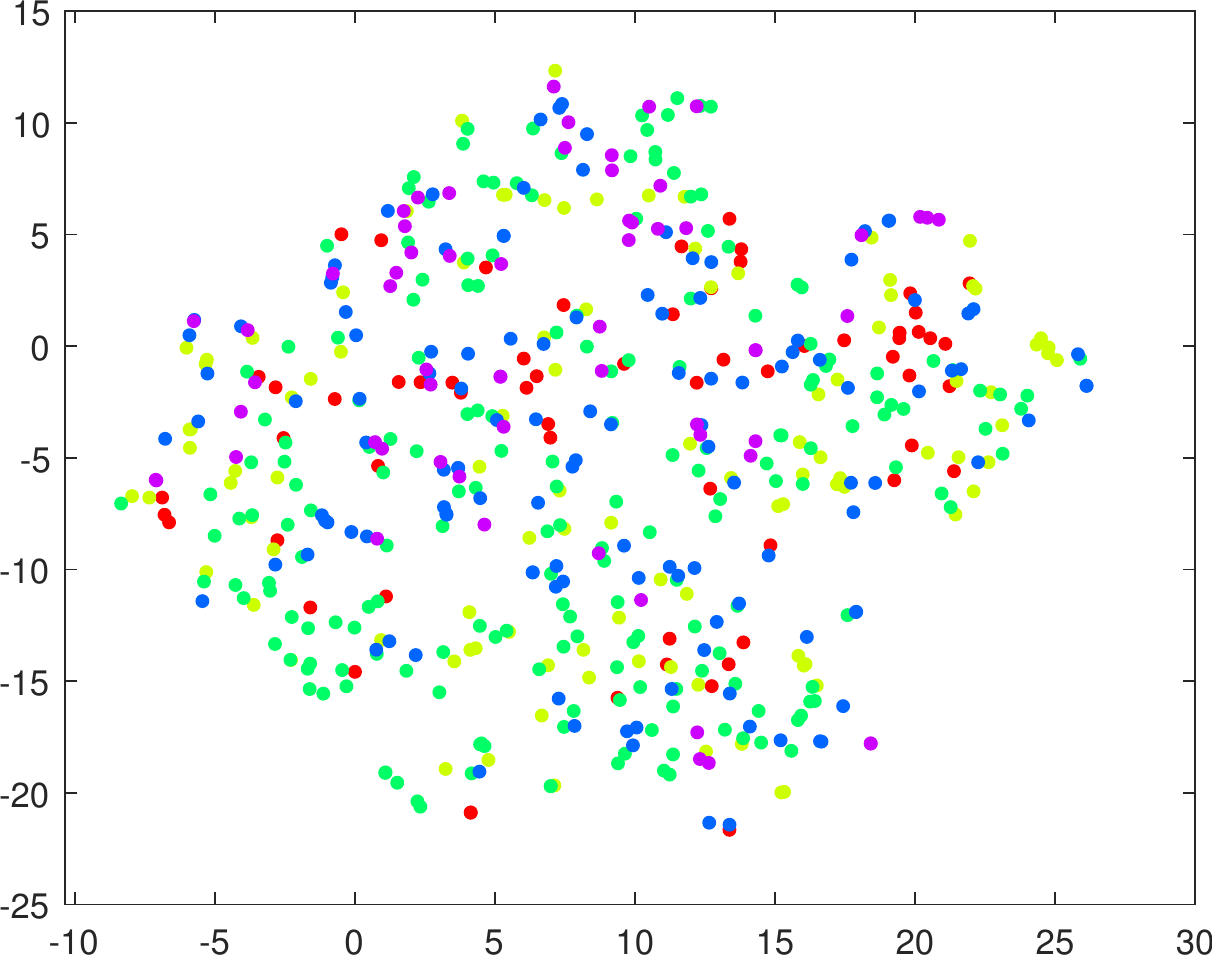}}
		\centerline{(b1) BBCSport(iter=1)}
	\end{minipage}
	\begin{minipage}{0.32\linewidth}
		\vspace{3pt}
		\centerline{\includegraphics[width=\textwidth]{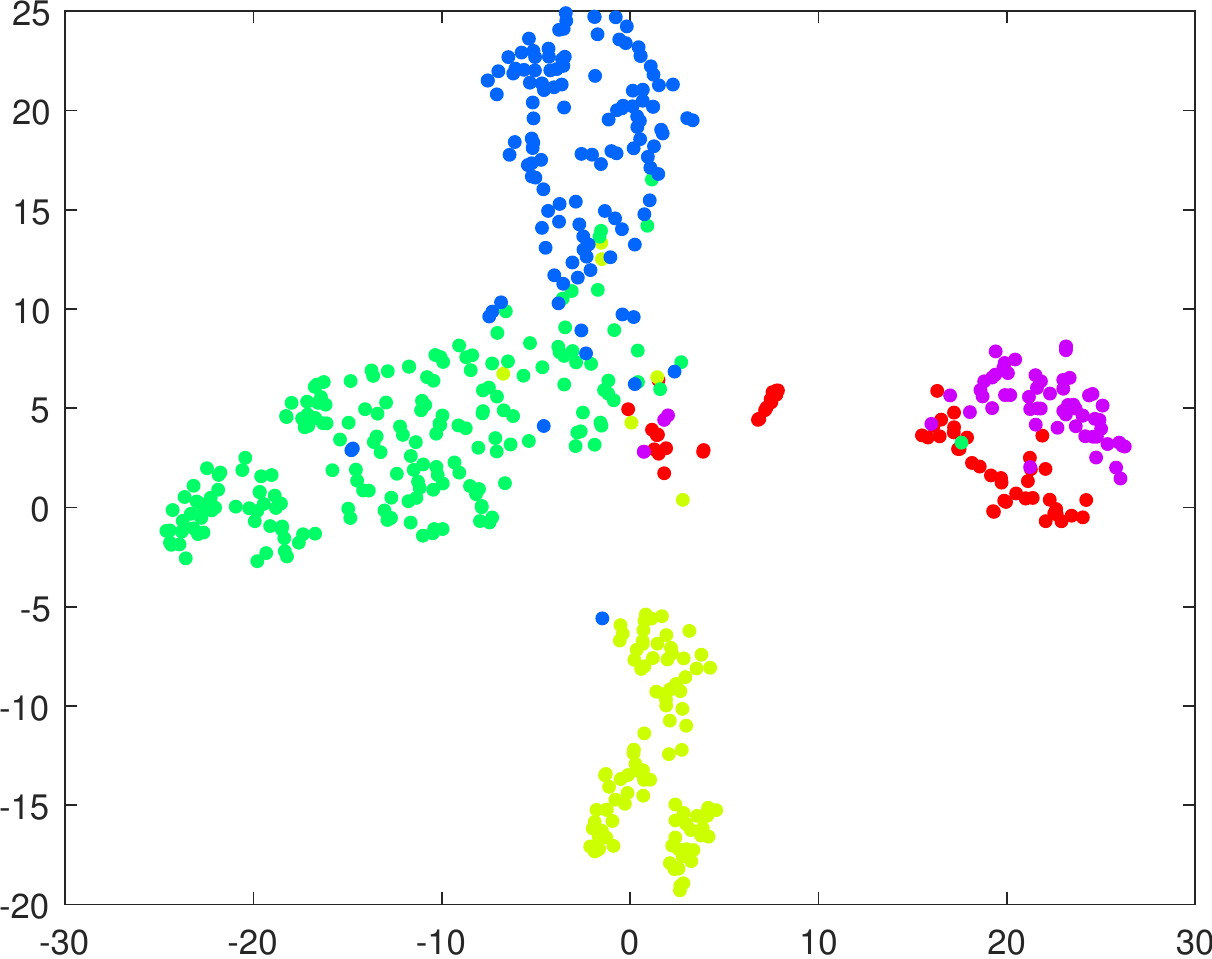}}
		\centerline{(b2) BBCSport(iter=10)}
	\end{minipage}
	\begin{minipage}{0.32\linewidth}
		\vspace{3pt}
		\centerline{\includegraphics[width=\textwidth]{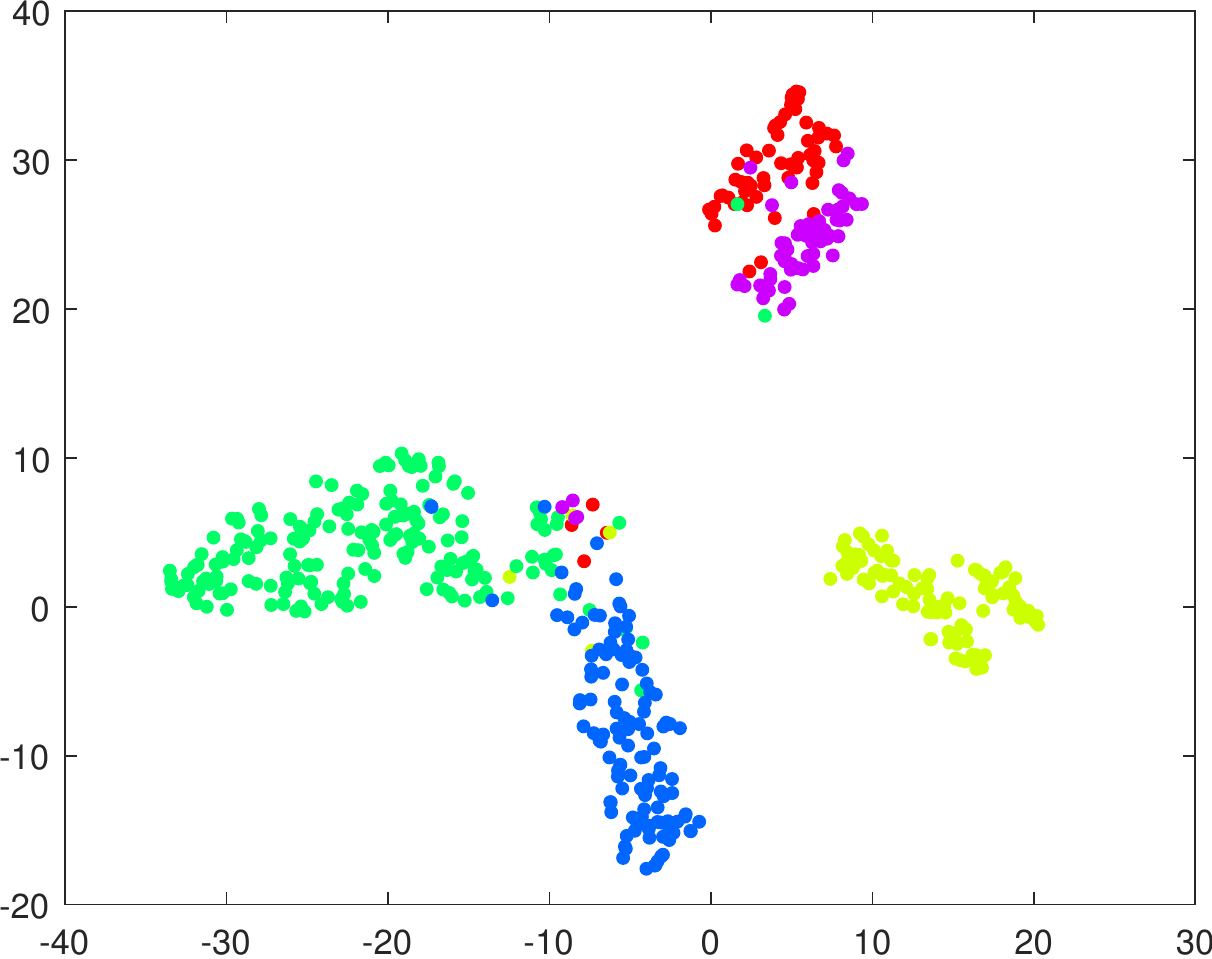}}
		\centerline{(b3) BBCSport(iter=60)}
	\end{minipage}	
	\caption{The proposed algorithm use t-SNE \cite{van2008visualizing} on BBC and BBCSport when iterations are 1, 10 and 60, The different colors indicate different classes for each dataset.}
	\label{fig:vis_iter}
\end{figure*} 

\subsection{Experiment results}

The ACC, NMI and Purity of the compared algorithm on the six benchmark datasets are displayed in Table \ref{tab:res_all}. The best are presented in bold. Table \ref{tab:res_increse} represents the incremental values of the three different metrics over the second-best method on six datasets, which is obtained from Tables \ref{tab:res_all}. From these tables, we have the following conclusions:
\begin{itemize}
	\item As shown in Table \ref{tab:res_increse}, on BBC data, the improvement is about $11.68\%$, $15.55\%$ and $3.47\%$ on ACC, NMI, Purity, respectively; on BBCSport data, the improvement is about $19.85\%$, $11.31\%$ and $17.46\%$ respectively. For NMI on Reuters and HW, although the performance is decreased by $2.28\%$ and $4.59\%$ compared to the second-best method, the difference is smaller. Overall, our proposed method (MVC-DMF-PA) outperforms the comparison baseline on six benchmarks.
	\item Comparing with the strong baseline DMVC and AwDMVC which also using the deep semi-NMF framework, we can find that we always achieve the best results. This means that our framework with post-fusion is more effective and robust for these datasets.
	\item Comparing with PMSC, which performs graph fusion first and then performs spectral clustering before late fusion, our method is more advantageous, further indicating that deep semi-NMF can extract more hidden useful information.
\end{itemize}

Overall, we have demonstrated the effectiveness of our method by the above experimental results. To summarize, our algorithm has the following advantages: $i)$ The quality of the base partition matrix is improved by obtaining the base partition matrix which containing deep and implicit information through deep semi-NMF framework. $ii)$ A late fusion approach is used to consider the locality of each single view, and the coefficients leading to the optimal clustering results are selected adaptively for each view to improve the accuracy of the clustering results.

\begin{figure*}
	\centering
	\begin{minipage}{0.16\linewidth}
		\vspace{3pt}
		\centerline{\includegraphics[width=\textwidth]{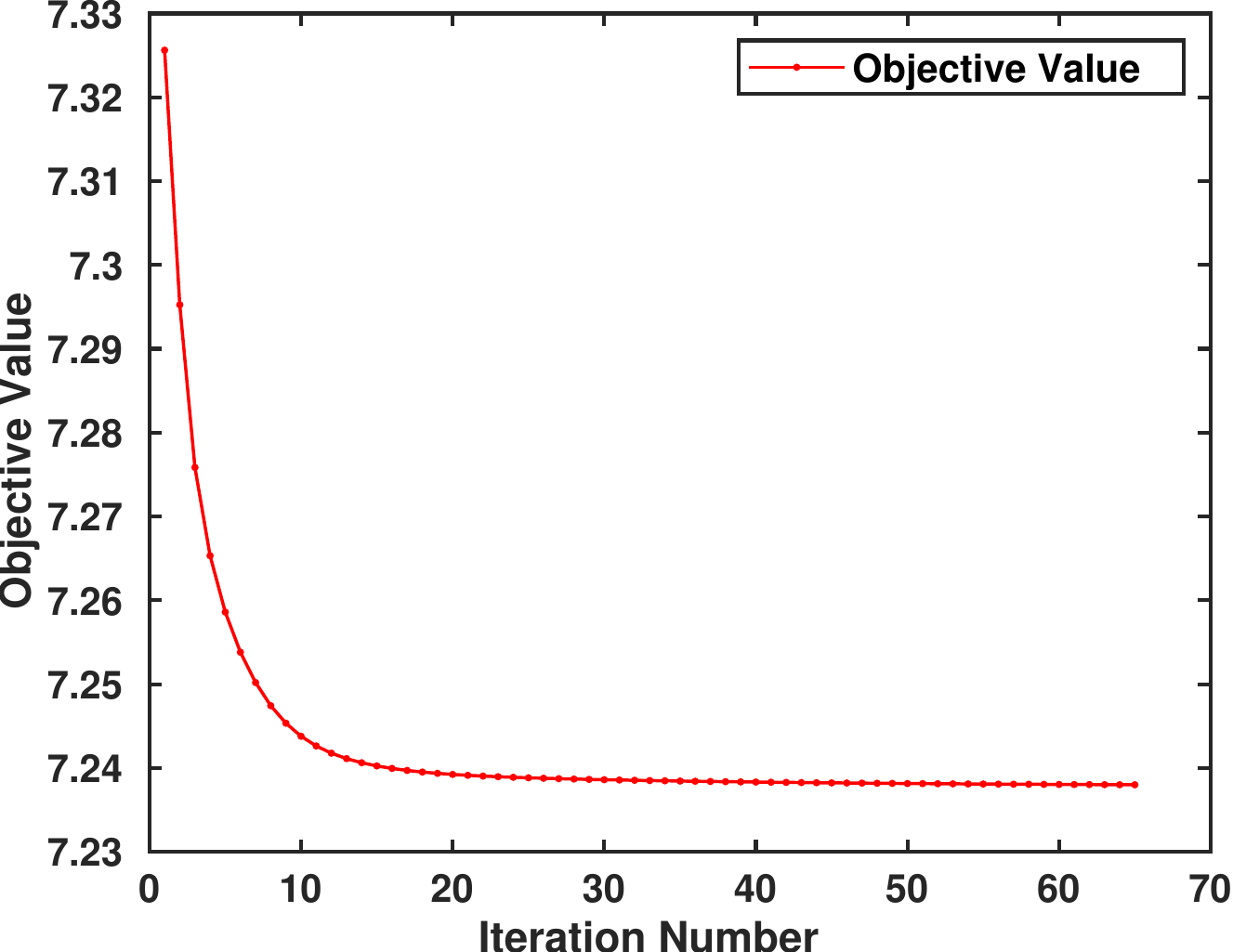}}
		\centerline{(a) BBC}
	\end{minipage}
	\begin{minipage}{0.16\linewidth}
		\vspace{3pt}
		\centerline{\includegraphics[width=\textwidth]{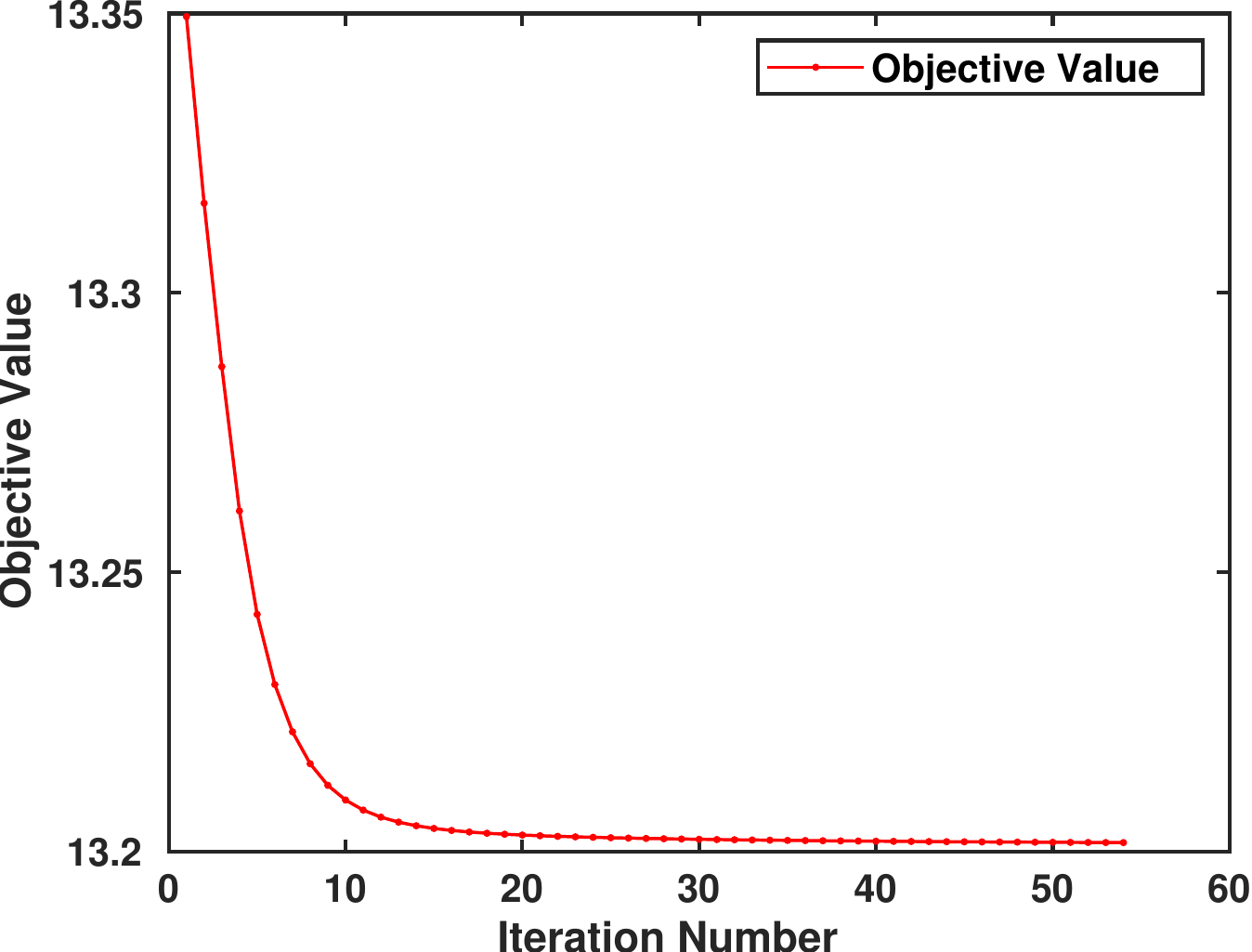}}
		\centerline{(b) BBCSport}
	\end{minipage}
	\begin{minipage}{0.16\linewidth}
		\vspace{3pt}
		\centerline{\includegraphics[width=\textwidth]{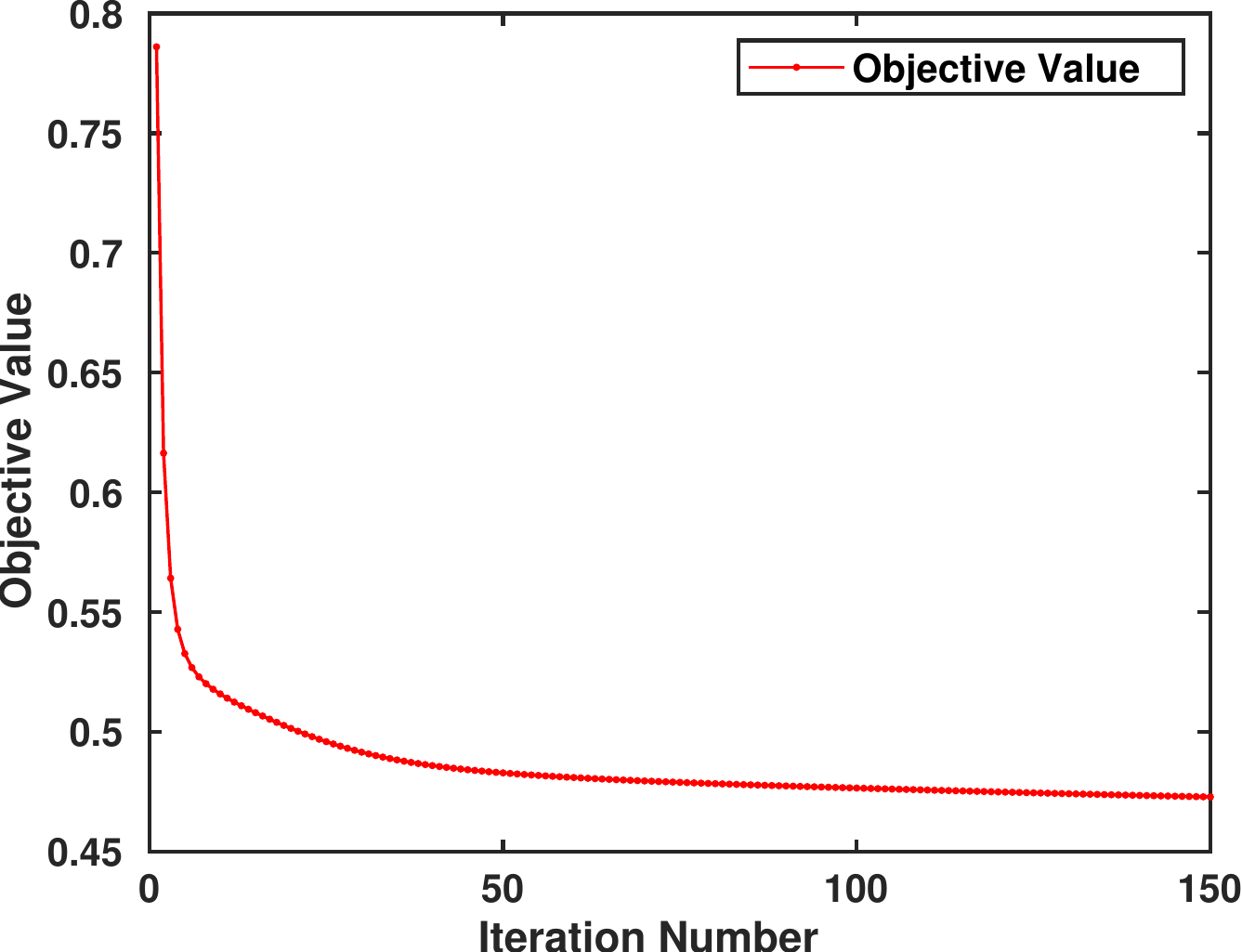}}
		\centerline{(c) MSRCV1}
	\end{minipage}	
	\begin{minipage}{0.16\linewidth}
		\vspace{3pt}
		\centerline{\includegraphics[width=\textwidth]{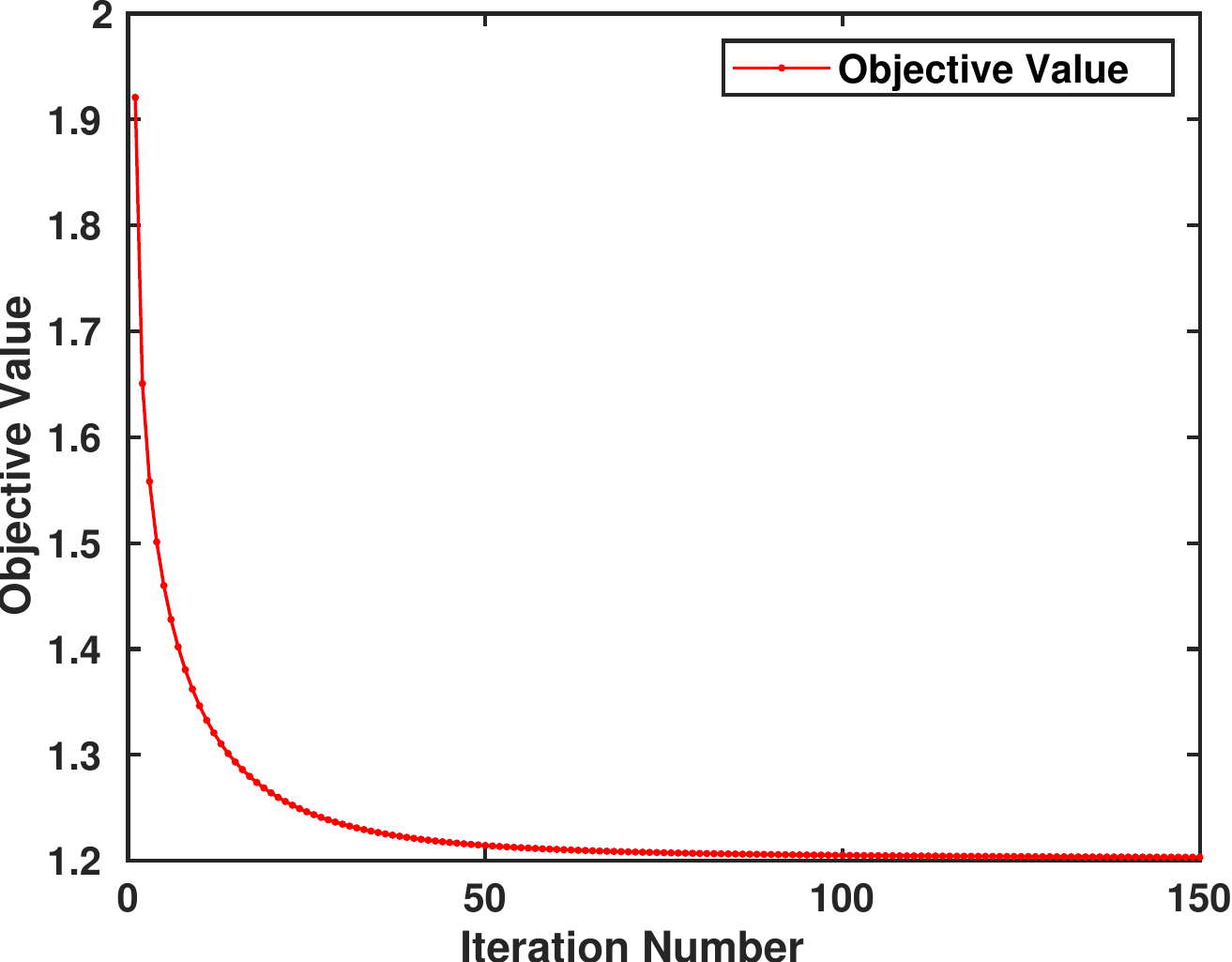}}
		\centerline{(d) Reuters}
	\end{minipage}
	\begin{minipage}{0.16\linewidth}
		\vspace{3pt}
		\centerline{\includegraphics[width=\textwidth]{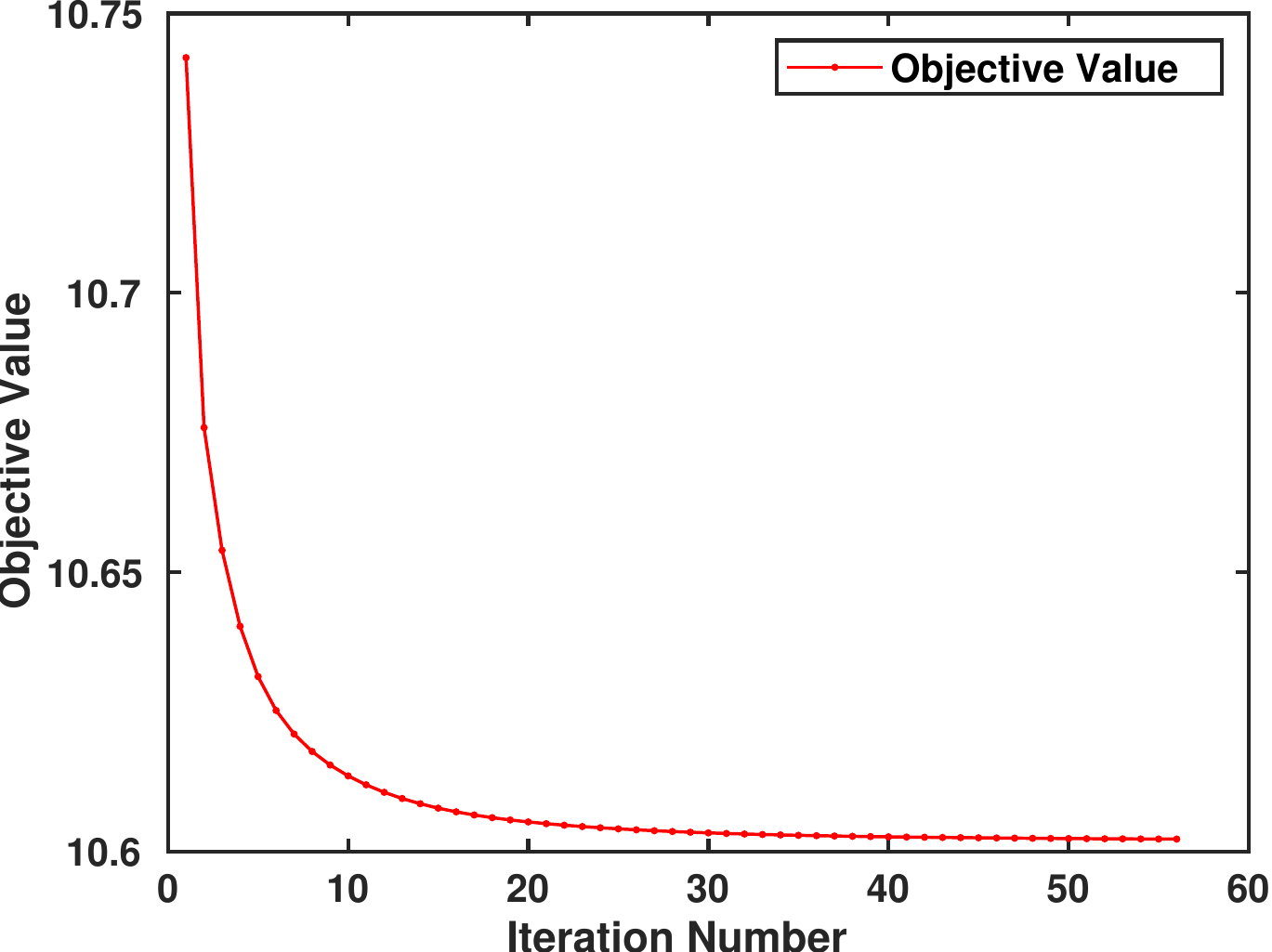}}
		\centerline{(d) Reuters}
	\end{minipage}
	\begin{minipage}{0.16\linewidth}
		\vspace{3pt}
		\centerline{\includegraphics[width=\textwidth]{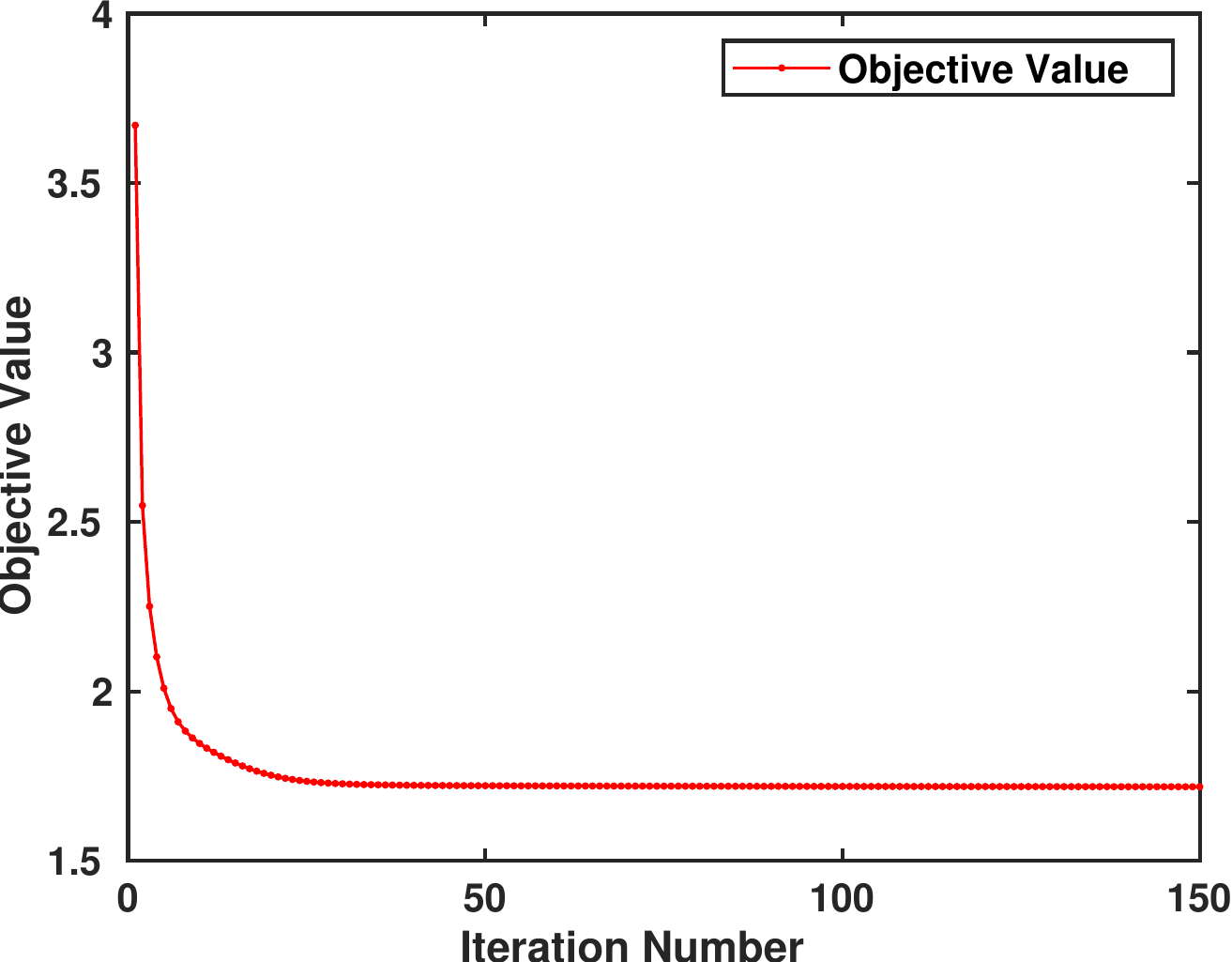}}
		\centerline{(e) HW}
	\end{minipage}
	\caption{The convergence of the proposed method on BBC, BBCSport, MSRCV1, ORL, Reuters and HW.}
	\label{fig:vis_obj}
\end{figure*} 
\begin{figure*}[t]
	\centering
	\begin{minipage}{0.16\linewidth}
		\vspace{3pt}
		\centerline{\includegraphics[width=\textwidth]{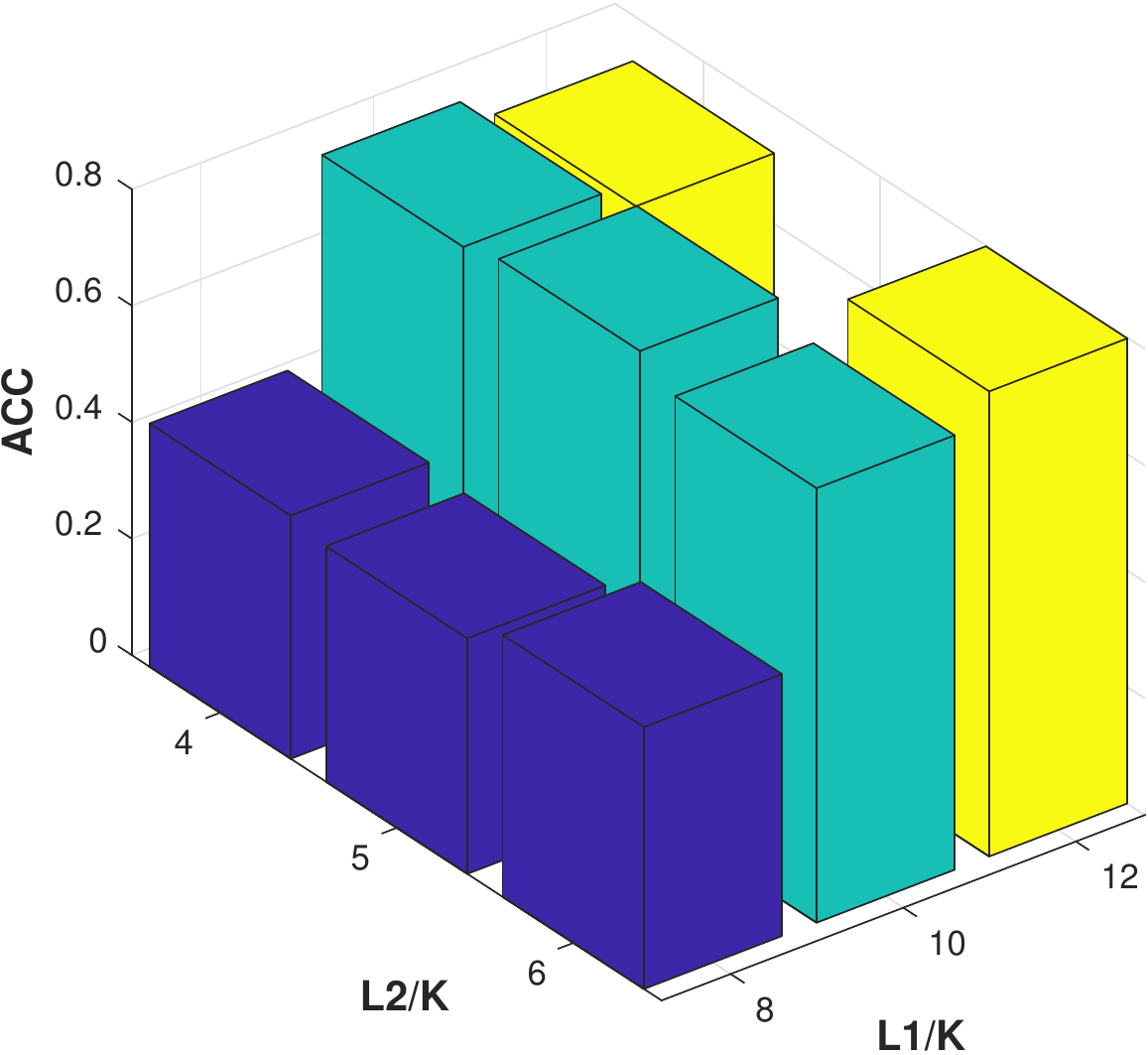}}
		\centerline{(a) BBC}
	\end{minipage}
	\begin{minipage}{0.16\linewidth}
		\vspace{3pt}
		\centerline{\includegraphics[width=\textwidth]{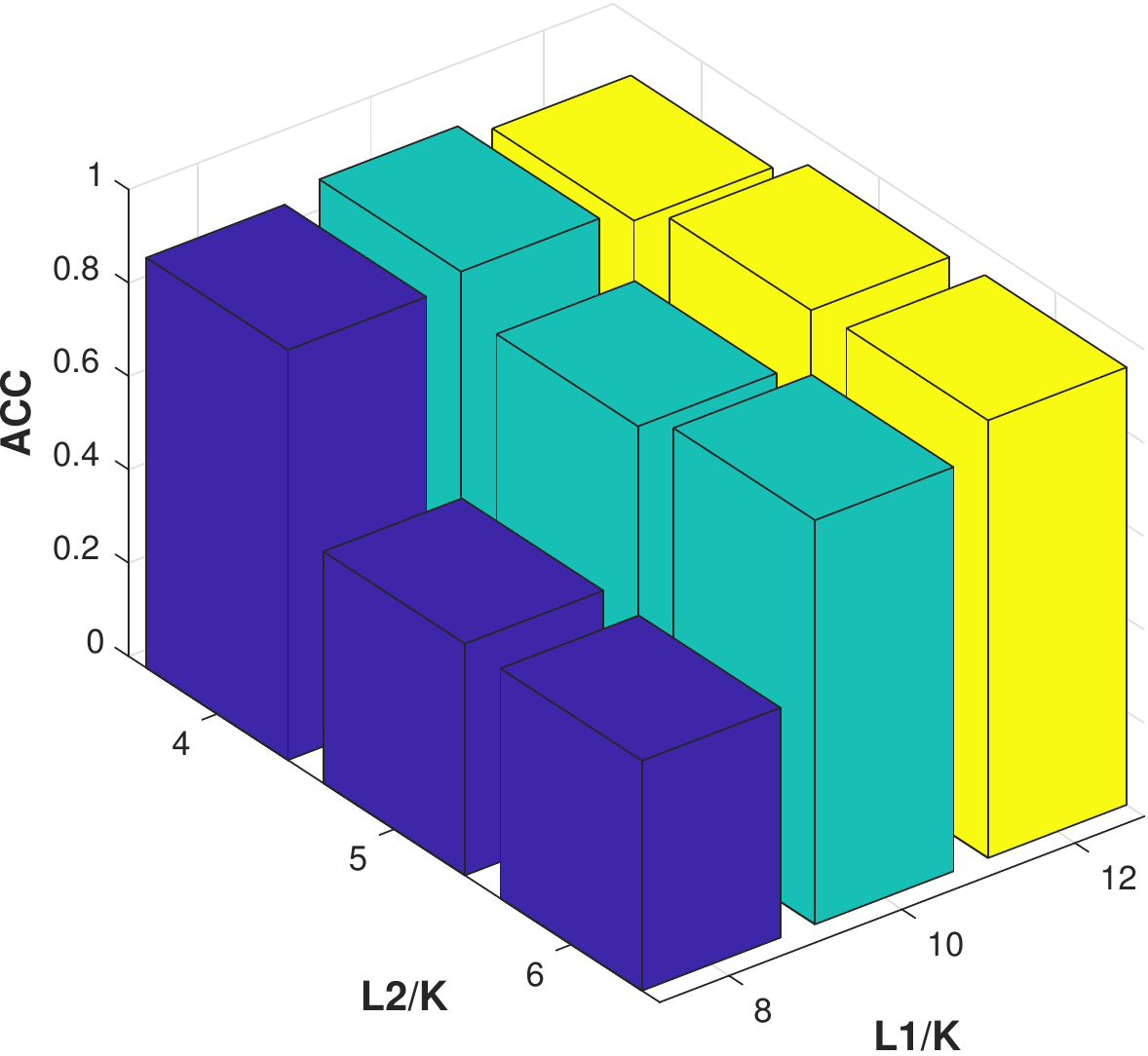}}
		\centerline{(b) BBCSport}
	\end{minipage}
	\begin{minipage}{0.16\linewidth}
		\vspace{3pt}
		\centerline{\includegraphics[width=\textwidth]{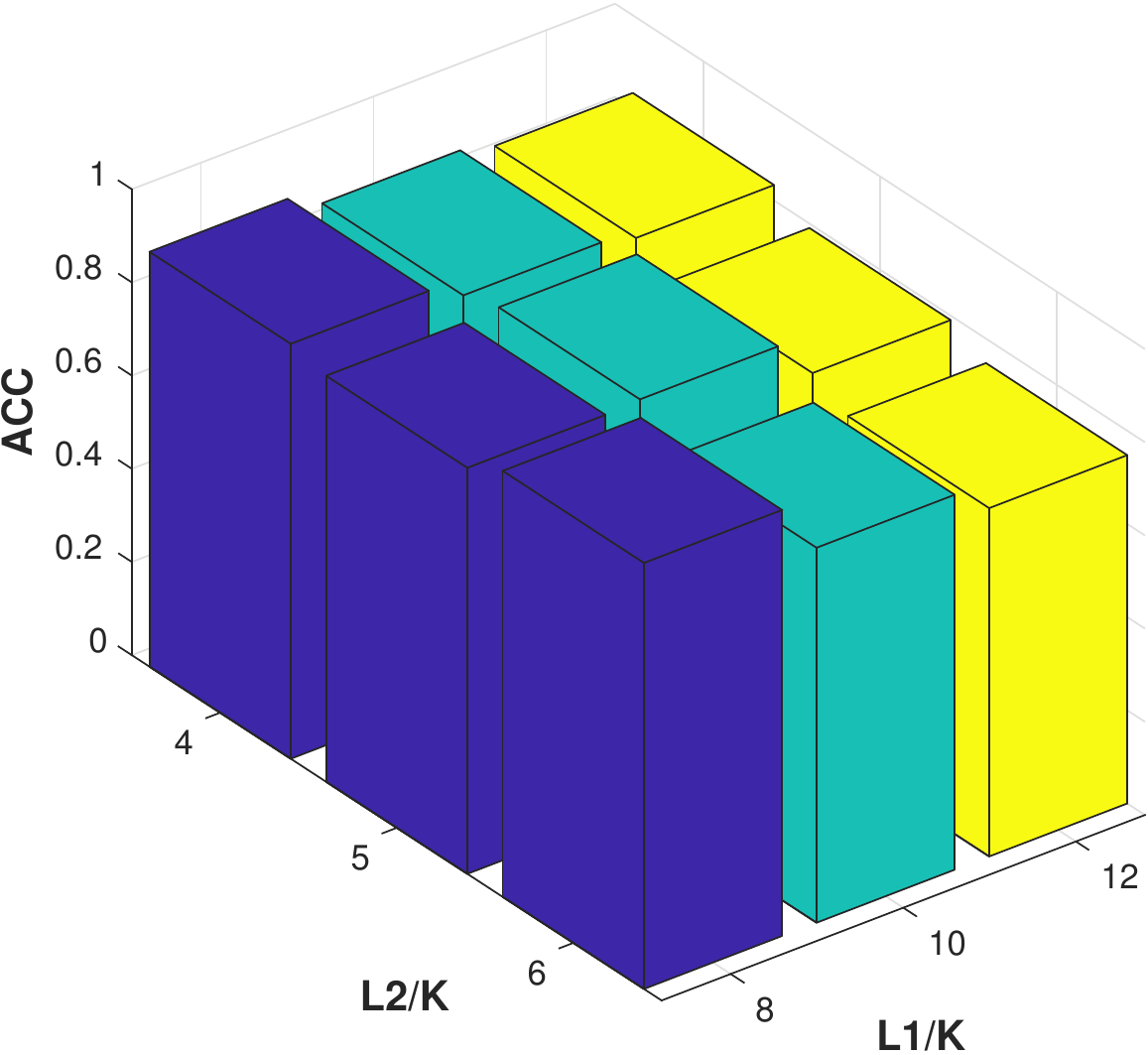}}
		\centerline{(c) MSRCV1}
	\end{minipage}
	\begin{minipage}{0.16\linewidth}
		\vspace{3pt}
		\centerline{\includegraphics[width=\textwidth]{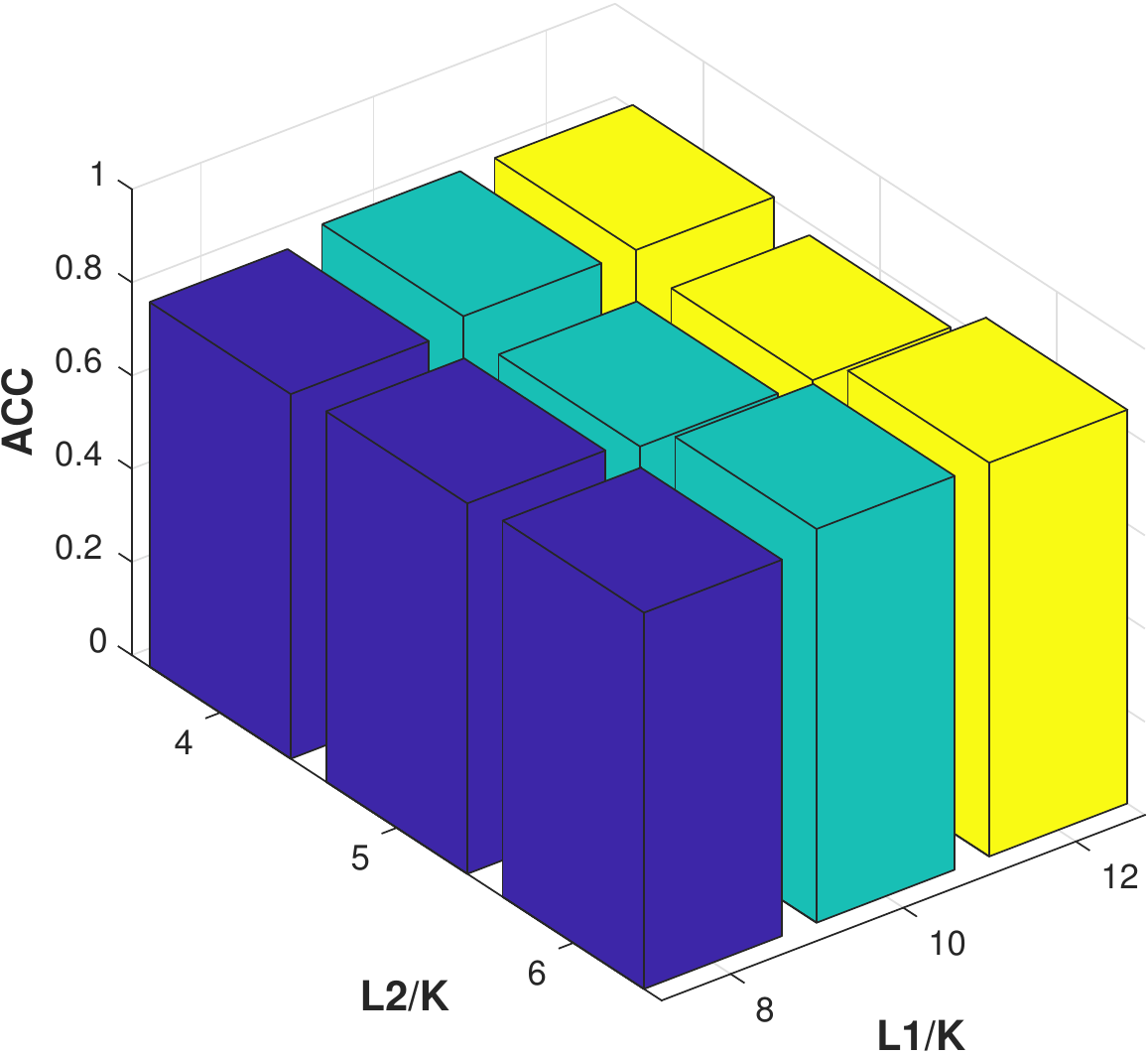}}
		\centerline{(d) ORL}
	\end{minipage}
	\begin{minipage}{0.16\linewidth}
		\vspace{3pt}
		\centerline{\includegraphics[width=\textwidth]{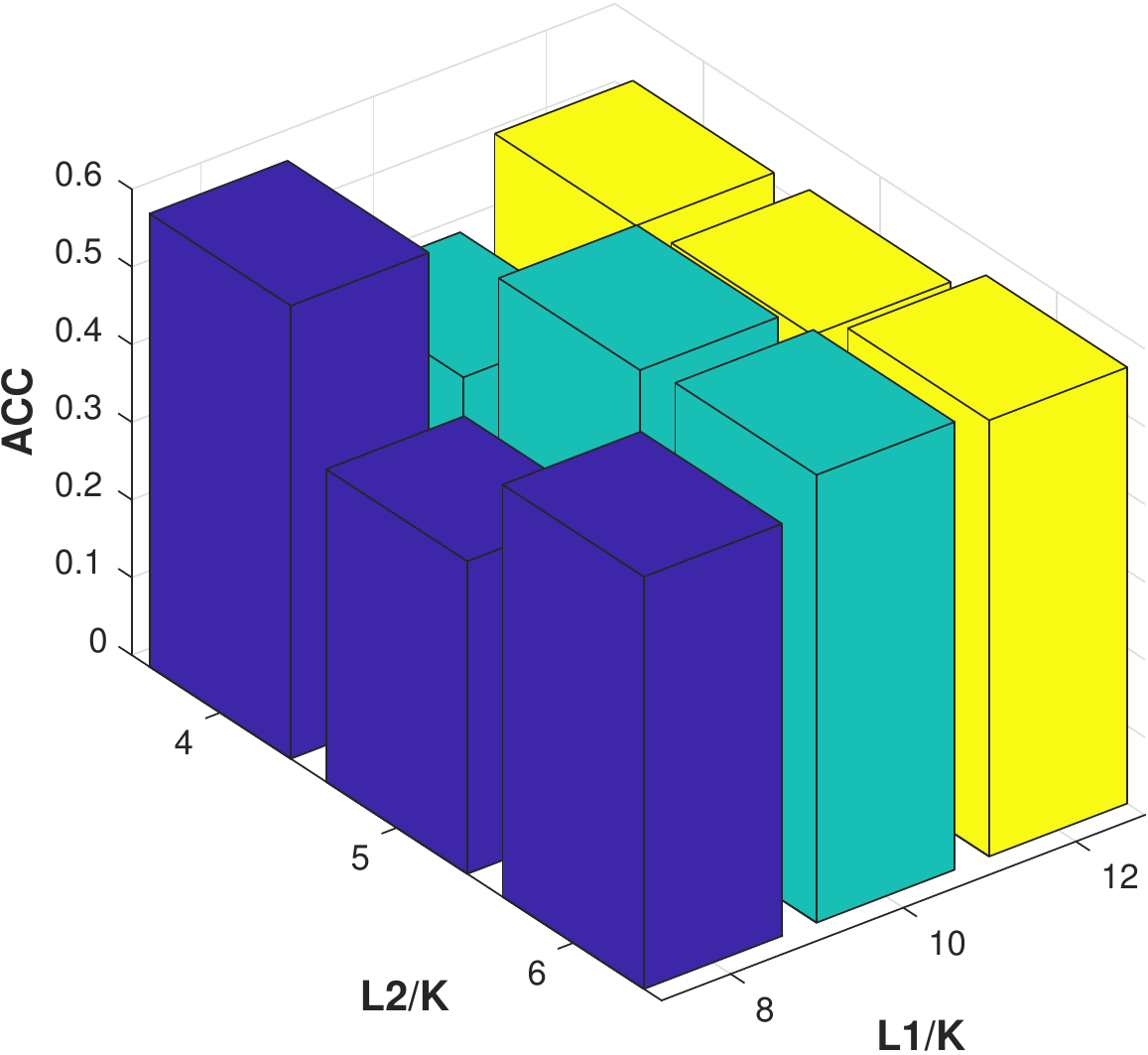}}
		\centerline{(e) Reuters}
	\end{minipage}	
	\begin{minipage}{0.16\linewidth}
		\vspace{3pt}
		\centerline{\includegraphics[width=\textwidth]{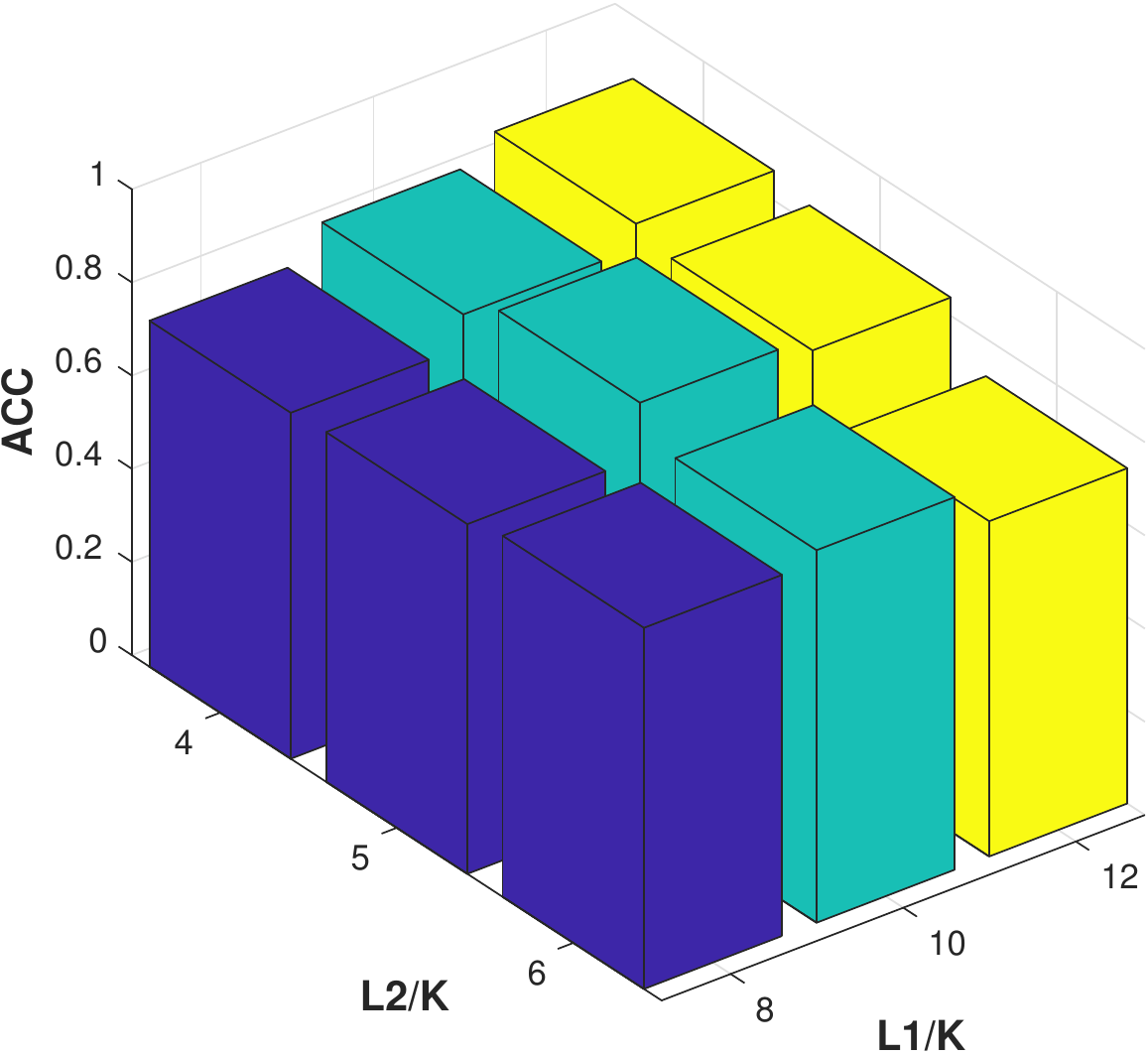}}
		\centerline{(f) HW}
	\end{minipage}
	\caption{The sensitivity of the proposed method with the variation of $l_1$ and $l_2$ in $p_3$ on on BBC, BBCSport, MSRCV1, ORL, Reuters and HW.}
	\label{fig:sen_l1l2}
\end{figure*}

\subsection{Visualization of the evolution of $\mathbf{H}$}

To demonstrate the effective of the consensus partition matrix $\mathbf{H}$, specifically, we evaluate the ACC of consensus partition $\mathbf{H}$ learned at each iteration, as shown in Figure \ref{fig:vis_iter}. We conduct the t-SNE algorithm \cite{van2008visualizing} on the consensus partition matrix $\mathbf{F}$ with different iterations, namely, $1^{st}$,$5^{th}$,$10^{th}$ and $20^{th}$ iteration. As the experimental results on Figure \ref{fig:vis_obj} shows, our algorithms quickly converge to a local minimum with less than 20 iterations.

Two examples of the evolution of consensus partition matrix $\mathbf{H}$ on BBC and BBCSport are demonstrated in Figure \ref{fig:vis_iter}. As Figure \ref{fig:vis_iter} shows, with the increasing number of iterations, the clustering structures of data become more significant and clearer than the old ones. These results clearly demonstrate the effectiveness of the learned consensus matrix $\mathbf{H}$ for clustering.

\subsection{Ablation study}

\begin{table}[htbp]
	\caption{Incremental values of three different metrics over the second best method on six datasets.}
	\label{tab:res_increse} 
	\centering
	\resizebox{\columnwidth}{9mm}{
		\begin{tabular}{ccccccc}
			\toprule
			Metric & BBC              & BBCSport         & MSRCV1 & ORL    & Reuters                        & HW                             \\
			\midrule
			ACC    & \textbf{11.68\%} & \textbf{19.85\%} & 1.90\% & 3.50\% & 6.40\%                         & 4.86\%                         \\
			NMI    & \textbf{15.55\%} & \textbf{11.31\%} & 3.47\% & 1.78\% & \textbf{-2.28\%} & \textbf{-4.59\%} \\
			PUR    & \textbf{3.47\%}  & \textbf{17.46\%} & 1.90\% & 3.75\% & 4.50\%                         & 4.33\%      \\
			\bottomrule                  
	\end{tabular}}
\end{table}

We record $l_1$, $l_2$ of parameter $p_3$ and $\gamma$ when the depth is three. As shown in Table \ref{tab:res_ablation} We also compare the values of ACC when the depth is one (the parameter are $p_1=\left[k\right]$, $\gamma$) and when the depth is two ($p_2=\left[l_2,k\right]$, $\gamma$). We find that the difference in performance between $p_1$ and $p_2$ is not significant, except for the dataset BBCSport where the performance advantage is doubled. The best performance is always achieved at $p_3$. In general, the results are better when the number of layers is deepened, which can be interpreted as deeper information is mined.

\begin{table}[H]
	\caption{ACC of different layers on six benchmark datasets.}
	\label{tab:res_ablation} 
	\centering
	\resizebox{\columnwidth}{9mm}{
		\begin{tabular}{ccccccc}
			\toprule
			$p$       & BBC             & BBCSport        & MSRCV1          & ORL             & Reuters         & HW              \\
			\midrule
			$[k]$       & 0.7387          & 0.4761          & 0.8286          & 0.7875          & 0.5825          & 0.7780          \\
			$[l_2,k]$    & 0.7343          & 0.8713          & 0.8238          & 0.7775          & 0.5258          & 0.7655          \\
			$[l_1,l_2,k]$ & \textbf{0.8102} & \textbf{0.9375} & \textbf{0.9143} & \textbf{0.8675} & \textbf{0.5908} & \textbf{0.8690}\\
			\bottomrule
	\end{tabular}}
\end{table}

\subsection{Convergence}

In Figs. \ref{fig:vis_obj} and \ref{fig:vis_iter}, we plot the change of target values during the iterations and the visualization of the clustering results. We can see that the target converges quickly in the first 10 iterations and it reaches convergence after 60 iterations, which can also be verified by Fig. \ref{fig:vis_iter}. Hence, it can be seen that our algorithm converges quickly. Moreover, the small fluctuations on the curve in Fig. \ref{fig:vis_obj} to can be explained by the iterative nature of our algorithm.

\subsection{Parameter sensitivity analysis}

In this section, we investigate the sensitivity of the parameters $p$ to the proposed method and explore how different values of the parameters will affect the performance of MVC-DMF-PA in multi-view clustering. We evaluate the sensitivity of the model to the parameters $l_1$, $l_2$ of $p_3$, $l_1$ and $l_2$ represent the dimension of the first and second layer when the feature matrix is decomposed in three layers. Fig. \ref{fig:sen_l1l2} shows the performance of MVC-DMF-PA for clustering with $l_1$ from $8k$ to $12k$ and $l_2$ from $4k$ to $6k$. From the figure, we can see that when we fix $l_1$, the performance mostly tends to decrease as $l_2$ increases, so most of the optimal values are obtained at $l_2=4k$. But the optimal value is uncertain for the value of $l_1$.

\section{Conclusion}\label{conclusion}

In this paper, we present MVC-DMF-PA, a Multi-View Clustering framework with Deep Matrix Factorization and Partition Alignment, to solve multi-view clustering problem about curse of dimensionality. First we use the depth matrix decomposition to obtain the base partitioning result for each view, and then fuse these partition matrix to approximate the common partition matrix. By alternatively updating the decomposition process as well as the late fusion process and the interaction of two processes, we can obtain a better common division result for clustering.Extensive experimental results on six benchmark show the effectiveness of our proposed method by comparing with 12 SOTA methods

\bibliographystyle{IEEEtran}
\bibliography{MVC-DMF-PA}

\begin{thebibliography}{10}
\providecommand{\url}[1]{#1}
\csname url@samestyle\endcsname
\providecommand{\newblock}{\relax}
\providecommand{\bibinfo}[2]{#2}
\providecommand{\BIBentrySTDinterwordspacing}{\spaceskip=0pt\relax}
\providecommand{\BIBentryALTinterwordstretchfactor}{4}
\providecommand{\BIBentryALTinterwordspacing}{\spaceskip=\fontdimen2\font plus
\BIBentryALTinterwordstretchfactor\fontdimen3\font minus
  \fontdimen4\font\relax}
\providecommand{\BIBforeignlanguage}[2]{{%
\expandafter\ifx\csname l@#1\endcsname\relax
\typeout{** WARNING: IEEEtran.bst: No hyphenation pattern has been}%
\typeout{** loaded for the language `#1'. Using the pattern for}%
\typeout{** the default language instead.}%
\else
\language=\csname l@#1\endcsname
\fi
#2}}
\providecommand{\BIBdecl}{\relax}
\BIBdecl

\bibitem{kumar_co-training_nodate}
A.~Kumar and H.~Daum{\'e}, ``A co-training approach for multi-view spectral
  clustering,'' in \emph{Proceedings of the 28th international conference on
  machine learning (ICML-11)}, 2011, pp. 393--400.

\bibitem{kumar_co-regularized_nodate}
A.~Kumar, P.~Rai, and H.~Daume, ``Co-regularized multi-view spectral
  clustering,'' \emph{Advances in neural information processing systems},
  vol.~24, pp. 1413--1421, 2011.

\bibitem{tan2020unsupervised}
J.~Tan, Y.~Shi, Z.~Yang, C.~Wen, and L.~Lin, ``Unsupervised multi-view
  clustering by squeezing hybrid knowledge from cross view and each view,''
  \emph{IEEE Transactions on Multimedia}, 2020.

\bibitem{li_multiple_nodate}
M.~Li, X.~Liu, L.~Wang, Y.~Dou, J.~Yin, and E.~Zhu, ``Multiple kernel
  clustering with local kernel alignment maximization,'' in \emph{Proceedings
  of the Twenty-Fifth International Joint Conference on Artificial
  Intelligence}, 2016, pp. 1704--1710.

\bibitem{huang2020robust}
B.~Huang, T.~Xu, S.~Jiang, Y.~Chen, and Y.~Bai, ``Robust visual tracking via
  constrained multi-kernel correlation filters,'' \emph{IEEE Transactions on
  Multimedia}, vol.~22, no.~11, pp. 2820--2832, 2020.

\bibitem{wang2020kernelized}
H.~Wang, Y.~Wang, Z.~Zhang, X.~Fu, L.~Zhuo, M.~Xu, and M.~Wang, ``Kernelized
  multiview subspace analysis by self-weighted learning,'' \emph{IEEE
  Transactions on Multimedia}, 2020.

\bibitem{chen2019jointly}
Y.~Chen, X.~Xiao, and Y.~Zhou, ``Jointly learning kernel representation tensor
  and affinity matrix for multi-view clustering,'' \emph{IEEE Transactions on
  Multimedia}, 2019.

\bibitem{wang2019multi}
S.~Wang, X.~Liu, E.~Zhu, C.~Tang, J.~Liu, J.~Hu, J.~Xia, and J.~Yin,
  ``Multi-view clustering via late fusion alignment maximization.'' in
  \emph{IJCAI}, 2019, pp. 3778--3784.

\bibitem{zhang_multilevel_2021}
H.~Zhang, D.~Wu, F.~Nie, R.~Wang, and X.~Li, ``Multilevel projections with
  adaptive neighbor graph for unsupervised multi-view feature selection,''
  \emph{Information Fusion}, vol.~70, pp. 129 -- 140, 2021.

\bibitem{chen_relaxed_2021}
M.-S. Chen, L.~Huang, C.-D. Wang, D.~Huang, and J.-H. Lai, ``Relaxed multi-view
  clustering in latent embedding space,'' \emph{Information Fusion}, vol.~68,
  pp. 8 -- 21, 2021.

\bibitem{zhou_subspace_2020}
S.~Zhou, E.~Zhu, X.~Liu, T.~Zheng, Q.~Liu, J.~Xia, and J.~Yin, ``Subspace
  segmentation-based robust multiple kernel clustering,'' \emph{Information
  Fusion}, vol.~53, pp. 145 -- 154, 2020.

\bibitem{wang2020learning}
B.~Wang, Y.~Hu, J.~Gao, Y.~Sun, F.~Ju, and B.~Yin, ``Learning adaptive
  neighborhood graph on grassmann manifolds for video/image-set subspace
  clustering,'' \emph{IEEE Transactions on Multimedia}, 2020.

\bibitem{zhou2007spectral}
D.~Zhou and C.~J. Burges, ``Spectral clustering and transductive learning with
  multiple views,'' in \emph{Proceedings of the 24th international conference
  on Machine learning}, 2007, pp. 1159--1166.

\bibitem{kang2020partition}
Z.~Kang, X.~Zhao, C.~Peng, H.~Zhu, J.~T. Zhou, X.~Peng, W.~Chen, and Z.~Xu,
  ``Partition level multiview subspace clustering,'' \emph{Neural Networks},
  vol. 122, pp. 279--288, 2020.

\bibitem{bickel2004multi}
S.~Bickel and T.~Scheffer, ``Multi-view clustering.'' in \emph{ICDM}, vol.~4,
  no. 2004.\hskip 1em plus 0.5em minus 0.4em\relax Citeseer, 2004, pp. 19--26.

\bibitem{ding2006orthogonal}
C.~Ding, T.~Li, W.~Peng, and H.~Park, ``Orthogonal nonnegative matrix
  t-factorizations for clustering,'' in \emph{Proceedings of the 12th ACM
  SIGKDD international conference on Knowledge discovery and data mining},
  2006, pp. 126--135.

\bibitem{zong2018multi}
L.~Zong, X.~Zhang, and X.~Liu, ``Multi-view clustering on unmapped data via
  constrained non-negative matrix factorization,'' \emph{Neural Networks}, vol.
  108, pp. 155--171, 2018.

\bibitem{wang2017diverse}
J.~Wang, F.~Tian, H.~Yu, C.~H. Liu, K.~Zhan, and X.~Wang, ``Diverse
  non-negative matrix factorization for multiview data representation,''
  \emph{IEEE transactions on cybernetics}, vol.~48, no.~9, pp. 2620--2632,
  2017.

\bibitem{yang2020uniform}
Z.~Yang, N.~Liang, W.~Yan, Z.~Li, and S.~Xie, ``Uniform distribution
  non-negative matrix factorization for multiview clustering,'' \emph{IEEE
  transactions on cybernetics}, 2020.

\bibitem{liu2020multi}
J.-W. Liu, Y.-F. Wang, R.-K. Lu, and X.-L. Luo, ``Multi-view non-negative
  matrix factorization discriminant learning via cross entropy loss,'' in
  \emph{2020 Chinese Control And Decision Conference (CCDC)}.\hskip 1em plus
  0.5em minus 0.4em\relax IEEE, 2020, pp. 3964--3971.

\bibitem{chen2019multiview}
F.~Chen, G.~Li, S.~Wang, and Z.~Pan, ``Multiview clustering via robust
  neighboring constraint nonnegative matrix factorization,'' \emph{Mathematical
  Problems in Engineering}, vol. 2019, 2019.

\bibitem{cai2010graph}
D.~Cai, X.~He, J.~Han, and T.~S. Huang, ``Graph regularized nonnegative matrix
  factorization for data representation,'' \emph{IEEE transactions on pattern
  analysis and machine intelligence}, vol.~33, no.~8, pp. 1548--1560, 2010.

\bibitem{zong2017multi}
L.~Zong, X.~Zhang, L.~Zhao, H.~Yu, and Q.~Zhao, ``Multi-view clustering via
  multi-manifold regularized non-negative matrix factorization,'' \emph{Neural
  Networks}, vol.~88, pp. 74--89, 2017.

\bibitem{wang2018multiview}
X.~Wang, T.~Zhang, and X.~Gao, ``Multiview clustering based on non-negative
  matrix factorization and pairwise measurements,'' \emph{IEEE transactions on
  cybernetics}, vol.~49, no.~9, pp. 3333--3346, 2018.

\bibitem{wu2018manifold}
B.~Wu, E.~Wang, Z.~Zhu, W.~Chen, and P.~Xiao, ``Manifold nmf with l21 norm for
  clustering,'' \emph{Neurocomputing}, vol. 273, pp. 78--88, 2018.

\bibitem{trigeorgis_deep_nodate}
G.~Trigeorgis, K.~Bousmalis, S.~Zafeiriou, and B.~Schuller, ``A deep semi-nmf
  model for learning hidden representations,'' in \emph{International
  Conference on Machine Learning}.\hskip 1em plus 0.5em minus 0.4em\relax PMLR,
  2014, pp. 1692--1700.

\bibitem{zhao2017multi}
H.~Zhao, Z.~Ding, and Y.~Fu, ``Multi-view clustering via deep matrix
  factorization,'' in \emph{Proceedings of the AAAI Conference on Artificial
  Intelligence}, vol.~31, no.~1, 2017.

\bibitem{huang_auto-weighted_2020}
S.~Huang, Z.~Kang, and Z.~Xu, ``Auto-weighted multi-view clustering via deep
  matrix decomposition,'' \emph{Pattern Recognition}, vol.~97, p. 107015, 2020.

\bibitem{ding2005equivalence}
C.~Ding, X.~He, and H.~D. Simon, ``On the equivalence of nonnegative matrix
  factorization and spectral clustering,'' in \emph{Proceedings of the 2005
  SIAM international conference on data mining}.\hskip 1em plus 0.5em minus
  0.4em\relax SIAM, 2005, pp. 606--610.

\bibitem{ding2008convex}
C.~H. Ding, T.~Li, and M.~I. Jordan, ``Convex and semi-nonnegative matrix
  factorizations,'' \emph{IEEE transactions on pattern analysis and machine
  intelligence}, vol.~32, no.~1, pp. 45--55, 2008.

\bibitem{ding_convex_2010}
C.~Ding, {Tao Li}, and M.~Jordan, ``Convex and {Semi}-{Nonnegative} {Matrix}
  {Factorizations},'' \emph{IEEE Transactions on Pattern Analysis and Machine
  Intelligence}, vol.~32, pp. 45--55, 2010.

\bibitem{tzortzis2012kernel}
G.~Tzortzis and A.~Likas, ``Kernel-based weighted multi-view clustering,'' in
  \emph{2012 IEEE 12th international conference on data mining}.\hskip 1em plus
  0.5em minus 0.4em\relax IEEE, 2012, pp. 675--684.

\bibitem{wang2019gmc}
H.~Wang, Y.~Yang, and B.~Liu, ``Gmc: Graph-based multi-view clustering,''
  \emph{IEEE Transactions on Knowledge and Data Engineering}, vol.~32, no.~6,
  pp. 1116--1129, 2019.

\bibitem{luo2018consistent}
S.~Luo, C.~Zhang, W.~Zhang, and X.~Cao, ``Consistent and specific multi-view
  subspace clustering,'' in \emph{Proceedings of the AAAI Conference on
  Artificial Intelligence}, vol.~32, no.~1, 2018.

\bibitem{wang2015feature}
Z.~Wang, X.~Kong, H.~Fu, M.~Li, and Y.~Zhang, ``Feature extraction via
  multi-view non-negative matrix factorization with local graph
  regularization,'' in \emph{2015 IEEE International conference on image
  processing (ICIP)}.\hskip 1em plus 0.5em minus 0.4em\relax IEEE, 2015, pp.
  3500--3504.

\bibitem{zhan2018adaptive}
K.~Zhan, J.~Shi, J.~Wang, H.~Wang, and Y.~Xie, ``Adaptive structure concept
  factorization for multiview clustering,'' \emph{Neural computation}, vol.~30,
  no.~4, pp. 1080--1103, 2018.

\bibitem{huang2018self}
S.~Huang, Z.~Kang, and Z.~Xu, ``Self-weighted multi-view clustering with soft
  capped norm,'' \emph{Knowledge-Based Systems}, vol. 158, pp. 1--8, 2018.

\bibitem{van2008visualizing}
L.~Van~der Maaten and G.~Hinton, ``Visualizing data using t-sne.''
  \emph{Journal of machine learning research}, vol.~9, 2008.

\end{thebibliography}

\end{document}